\documentclass[lettersize,journal]{IEEEtran}

\usepackage[nocompress]{cite}
\usepackage{graphicx} 
\usepackage{float} 
\usepackage{epstopdf}
\usepackage{color}
\usepackage{subfigure}
\usepackage{booktabs}
\usepackage{threeparttable}
\usepackage{multirow}
\usepackage{array}
\usepackage{algorithm}
\usepackage{algorithmicx}
\usepackage{algpseudocode}

\usepackage{url}
\usepackage{cite}
\usepackage{amsmath}
\usepackage{amssymb}
\usepackage{cases}
\usepackage{soul}
\usepackage{color}
\usepackage[hidelinks=true]{hyperref}

\newtheorem{lemma}{\textbf{Lemma}}
\newtheorem{remark}{\textbf{Remark}}

\newcommand{\lemmawithcite}[1]{%
  \refstepcounter{lemma}%
  \textbf{\textit{Lemma}} \textit{\thelemma}\cite{#1}\textit{:}%
}

\begin{document}

\title{Multi-view Spectral Clustering on the Grassmannian Manifold With Hypergraph Representation}

\author{Murong Yang, Shihui Ying,~\IEEEmembership{Member,~IEEE,} Xin-Jian Xu, Yue Gao,~\IEEEmembership{Senior Member,~IEEE}
\thanks{This work was supported by the National Key R\&D Program of China under Grant 2021YFA1003004 and Natural Science Foundation of China under Grant 12071281. (\it{Corresponding authors: Xin-Jian Xu; Yue Gao.})}
\thanks{Murong Yang is with the Department of Mathematics, College of Sciences, Shanghai University, Shanghai 200444, China (e-mail: mryang@shu.edu.cn).}
\thanks{Shihui Ying is with the
Shanghai Institute of Applied Mathematics and Mechanics, Shanghai University, Shanghai 200072, China and also with the School of Mechanics and Engineering Science, Shanghai University, Shanghai 200072, China
 (e-mail: shying@shu.edu.cn).}
\thanks{Xin-Jian Xu is with the Qian Weichang College, Shanghai University, Shanghai 200444, China (e-mail: xinjxu@shu.edu.cn).}
\thanks{Yue Gao is with BNRist, KLISS, School of Software, Tsinghua University, Beijing 100084, China and also with THUIBCS, Tsinghua University, Beijing 100084, China
(e-mail: gaoyue@tsinghua.edu.cn).}
}

\markboth{Journal of \LaTeX\ Class Files,~Vol.~14, No.~8, August~2021}%
{Shell \MakeLowercase{\textit{et al.}}: A Sample Article Using IEEEtran.cls for IEEE Journals}

\IEEEpubid{0000--0000/00\$00.00~\copyright~2021 IEEE}

\maketitle

\begin{abstract}
Graph-based multi-view spectral clustering methods have achieved notable progress recently, yet they often fall short in either oversimplifying pairwise relationships or struggling with inefficient spectral decompositions in high-dimensional Euclidean spaces. In this paper, we introduce a novel approach that begins to generate hypergraphs  by leveraging sparse representation learning from data points. Based on the generated hypergraph, we propose an optimization function with orthogonality constraints for multi-view hypergraph spectral clustering, which incorporates spectral clustering for each view and ensures consistency across different views.
In Euclidean space, solving the orthogonality-constrained optimization problem may yield local maxima and approximation errors.
Innovately, we transform this problem  into an unconstrained form on the Grassmannian manifold. Finally, we devise an alternating iterative Riemannian optimization algorithm to solve the problem.
To validate the effectiveness of the proposed algorithm, we test it on four real-world multi-view datasets and compare its performance with seven state-of-the-art multi-view clustering algorithms. The experimental results demonstrate that our method outperforms the baselines in terms of clustering performance due to its superior low-dimensional and resilient feature representation.
\end{abstract}

\begin{IEEEkeywords}
Hypergraph representation, spectral clustering, Grassmannian manifold, multi-view learning.
\end{IEEEkeywords}

\section{Introduction}
\IEEEPARstart{M}{ulti-view} data is prevalent in real-world applications due to the rapid development of multimedia technology. For instance, an image can be characterized by various features such as its color, texture, and shape. Similarly, a webpage can be described the documents it contains  as well as the hyperlinks associated with it.  This multi-view representation of data offers a wealth of information, revealing underlying structures more comprehensively than traditional single-view data. To leverage the  full potential information from each view and improve task performance, multi-view learning method has been developed \cite{8731740}.
With the scarcity of prior knowledge and the high costs of manual labeling in data collection, the study of multi-view clustering has become increasingly important and has attracted significant attention in recent years \cite{8502831,9395530}.

Graph-based methods are highly useful for multi-view clustering, as they effectively capture pairwise relationships between samples \cite{8662703}. A common approach learns a unified graph from individual view-specific graphs for clustering. For instance, Liang et al. \cite{8970909} introduced a consistency graph learning framework (CGL) with a regularization term to reduce cross-view inconsistencies. Wang et al. \cite{8662703} developed a graph-based multi-view clustering framework (GMC) that automatically assigns weights to each view to derive a unified graph matrix. Similarly, Chen et al. \cite{2020MultiChen} proposed a multi-view clustering approach in latent embedding space (MCLES) to capture global structure. Additionally, Huang et al. \cite{2021Measuring} presented a multi-view graph clustering method based on consistency and diversity (CDMGC) in a unified framework. With the rich spectral information of graph Laplacian, spectral clustering has become popular for multi-view clustering. For example, Kumar et al. \cite{NIPS2011_31839b03} introduced CoRegSC, a co-regularized spectral clustering method for multi-view data. And  Liu et al. \cite{Zong} proposed a weighted multi-view spectral clustering algorithm (WMSC) that employs spectral perturbation to simultaneously determine the weight of each view.  However, these methods capture only pairwise relationships, which limits their effectiveness in clustering real-world data with higher-order relationships.

\IEEEpubidadjcol

Fortunately, hypergraphs have proven effective for capturing higher-order relationships in data. Zhou et al. \cite{2006Learning} introduced a hypergraph Laplacian matrix to facilitate hypergraph partitioning, which lays the foundation for hypergraph learning. Building on this, Ma et al. \cite{9535255} proposed pLapHGNN, a p-Laplacian-based model that enhances representation learning by leveraging manifold geometry. More recently, Ma et al. \cite{10336546} introduced HGRec which boosts recommendation performance through automatic hypergraph generation and sparse optimization.
For multi-view data, Gao et al. \cite{6200340} constructed a multi-hypergraph framework to improve 3-D object retrieval and recognition. And Liu et al. \cite{7993002} proposed a transductive learning method for multi-hypergraphs by integrating hypergraph regularization for incomplete data handling.  Nevertheless, how to generatevalid hypergraphs as the input of these frameworks remains challenging. Current hypergraph generation approaches are categorized as explicit and implicit.
Explicit approaches utilize attributes or network structures. Attribute-based methods connect vertices with sharing attributes (e.g., color or shape \cite{Huang_2015_CVPR}) but are limited by attribute availability. Network-based methods use connectivity information, which often converts networks into simple graphs or leverages k-hop neighbors \cite{10.1145/3308558.3313635}. While these methods leverage attributes or connections, they may overlook certain potential relationships.
Implicit approaches can explore underlying relationships. A common approach is the distance-based methods that connect vertices by proximity in feature space (e.g., k-NN, $\epsilon$-ball). They are sensitive to noise and lack global exploration \cite{7064739}. Alternatively, sparse representation methods reconstruct features to establish connections, which reveal underlying structures. Wang et al. \cite{7064739} introduced this for $l_1$-hypergraph generation, later enhanced by Liu et al. \cite{2016Elastic} with Elastic-hypergraph.  Then Jin et al. \cite{2019Robust} further proposed a robust $l_2$-hypergraph model for nonlinear data. In this paper, we adopt the sparse representation approach for its robust representation and noise resilience.

On the other hand, subspace learning enhances clustering by assuming that data points originate from a combination of multiple low-dimensional subspaces \cite{8590732}. Manifolds, as a variant of subspaces, provides more adaptability for handling nonlinear data structures. By modeling on manifolds, it enhances robustness to noise and reduces computational complexity for high-dimensional data.  Hu et al. \cite{9859633} proposed a hypergraph-based multi-view clustering model on the Stiefel manifold, which combines views through manifold averaging. However, as the Stiefel manifold depends on the basis choice in subspaces, it lacks uniqueness in solutions. The Grassmann manifold that represents all $k$-dimensional linear subspaces overcomes this limitation. Some studies have directly applied graph-based spectral clustering on the Grassmannian manifold. For example, Wang et al. \cite{8099818} refined the sparse spectral clustering objective through Grassmannian optimization to derive latent embeddings. And Pasadakis et al. \cite{Pasadakis} reformulated the $k$-way $p$-spectral clustering as an unconstrained Grassmannian optimization. The Grassmannian manifold inherently provides a geometric framework for eigenspace subspaces with orthogonal constraints. However, hypergraph-based models or multi-view clustering models on the Grassmannian manifold remain underexplored.

In this paper, we propose a novel framework named multi-view hypergraph spectral clustering on the Grassmannian manifold (MHSCG). We first generate hypergraphs through sparse representation learning. Then we develop an optimization function for multi-view spectral clustering and transform this problem with orthogonality constraints on the Grassmannian manifold. Finally, we design an alternating iterative Riemannian optimization algorithm to solve it. The key contributions are outlined below:

\begin{itemize}
    \item[$\bullet$]
    We generate hypergraphs based on data similarity matrices derived from sparse representation learning, which effectively captures potential higher-order relationships and ensures robustness. 

    \item[$\bullet$]
    Based on the generated hypergraphs, we propose a multi-view spectral clustering function that minimizes hyperedge cut ratios by using the top $k$ Laplacian eigenvectors and ensures consistency by minimizing discrepancies across views simultaneously.

    \item[$\bullet$]
    In Euclidean space, finding orthogonal eigenvectors can result in local maxima. We address this issue by transforming the constrained problem into an unconstrained one on the Grassmannian manifold. 

    \item[$\bullet$]
     To solve this problem, we design a Riemannian optimization algorithm with an alternating iterative technique. We adaptively update the weighted parameters, which removes the need for manual tuning.

    \item[$\bullet$] The experimental results  demonstrate that the proposed algorithm outperforms seven baselines in terms of four clustering metrics.
\end{itemize}

The remaining sections are organized as follows: Section \ref{sr} reviews related works. Section \ref{s2} describes hypergraph generation and the optimization problem for multi-view hypergraph spectral clustering on the Grassmannian manifold. Section \ref{s3} presents the proposed Riemannian optimization algorithm. Section \ref{s4} represents the experimental results, and Section \ref{s5} concludes the paper.

\section{Related Works}\label{sr}

In this section, we provide a brief overview of related works, including multi-view spectral clustering, hypergraph learning and subspace learning on the Grassmannian manifolds.

\subsection{Multi-view Spectral Clustering  }

Traditional multi-view spectral clustering relies on constructing graphs for each view and utilizes the graph Laplacian to capture significant structural information for enhanced data representation across views \cite{2011Multiview,2018Multiview}. Within this framework, the multi-view spectral embedding (MSE) method identifies a shared low-dimensional embedding space, effectively harnessing the complementary information inherent in each view \cite{2011Multiview}.
Beyond this approach, the co-training spectral clustering technique iteratively refines the eigenvector matrix of one view by integrating clustering information from another, which strives for refined clustering results through mutual reinforcement \cite{2011A}. Appice et al. \cite{2016A} developed a method called co-training based trace distance-based clustering (CoTraDiC), which implements a co-training mechanism to derive multiple trace profiles and enrich the clustering process with cross-view learning.
In addition,  Kumar et al. \cite{NIPS2011_31839b03} introduced CoRegSC with co-regularization terms that enforce consistency among multiple views within the objective function of traditional spectral clustering and align view-specific representations. Based on this, Huang et al. \cite{6805155} further advanced it with a co-learned multi-view spectral clustering model that relaxes typical orthogonality constraints. This model opts for an alternative solution space that omits rigid eigenvector constraints. To Address the challenges in multi-view nonconvex sparse spectral clustering (SSC), Lu et al. \cite{7451227} proposed a convex relaxation technique by constructing a convex hull of fixed-rank projection matrices, which provides a robust alternative to conventional nonconvex approaches.

\subsection{Hypergraph Learning}

Hypergraph learning has garnered significant attention for its effectiveness in a wide range of applications, including image retrieval and social network analysis \cite{2010Image,2018Learning}. In the spectral analysis of hypergraphs, defining the Laplacian matrix directly on the hypergraph is a commonly employed method to ensure the retention of complex relational information inherent to hypergraphs. Zhou et al. \cite{2006Learning} developed an N-cut algorithm based on the normalized hypergraph Laplacian, which has become a foundational approach for hypergraph embedding, clustering, and transductive inference. This method has since found extensive application, particularly in the field of image recognition \cite{Xueqi2018Hypergraph}. Building on these concepts, Carletti et al. \cite{2019Random} introduced a hypergraph random walk algorithm that leverages the generalized hypergraph Laplacian operator, which extends its utility in modeling complex relationships.
In the realm of deep hypergraph learning, Feng et al. \cite{2019HypergraphFeng} proposed a hypergraph neural network (HGNN) framework, which integrates Zhou's hypergraph Laplacian into traditional graph convolutional neural networks, which enables the model to handle high-order relationships present in hypergraphs effectively. Further advancing this idea, Bai et al. \cite{2021Hypergraph} combined hypergraph attention mechanisms with hypergraph convolution to create a robust learning approach for high-order graph-structured data and improve the representation power in tasks involving intricate dependencies.

\subsection{Subspace Learning on the Grassmannian Manifold}


Subspace learning offers the advantage of reducing high-dimensional data into compact, interpretable representations that reveal underlying patterns. In recent years, interest in subspace learning has surged significantly, with particular attention to techniques on the Grassmannian manifold. The geometric structure of the Grassmannian manifold offers a robust framework to represent data.  It enables data points with similar patterns to be clustered within a common subspace and enhances the interpretability of clustering results. Edelman et al. \cite{1998The} expanded the applicability of Grassmannian manifold techniques by introducing novel Newton and conjugate gradient algorithm. This algorithm is specifically designed for optimization on the Grassmannian and Stiefel manifolds, which are efficient and adaptable to complex data structures. Building on this foundation, Ham et al. \cite{2008Grassmann} proposed a discriminant learning framework that incorporates distance analysis and positive semidefinite kernels on the Grassmannian manifold to enable more refined pattern separability.
Further research has leveraged Riemannian optimization on the Grassmannian manifold to address complex clustering problems. For example, Wang et al. \cite{8099818} employed Riemannian optimization techniques to tackle sparse spectral clustering on the Grassmannian manifold, which achieves promising results in sparse data settings by capturing essential structures. Additionally, Pasadakis et al. \cite{Pasadakis} refined this approach by optimizing objective functions with orthogonal constraints. It effectively aligns data points with subspaces on the Grassmannian manifold and further reinforce the clustering process.

\section{Multi-view Hypergraph Spectral Clustering on the Grassmannian Manifold}\label{s2}

In this section, we begin by generating a hypergraph for each view through sparse representation learning. Then we integrate them as inputs into a optimization problem for multi-view spectral clustering. Finally, we transform this problem with orthogonal constraints on the Grassmannian manifold. 


\subsection{Hypergraph Generation}
Given a dataset with its feature matrix represented as $\mathbf{X}=[\mathbf{x}_1,\mathbf{x}_2,\ldots,\mathbf{x}_n]^T\in \mathbb{R}^{n\times d}$. Each data point corresponds to a vertex in the finite set $V=\{v_1,v_2,\ldots,v_n\}$. A hypergraph $G=(V, E)$ is constructed, where $E=\{e_1,e_2,\ldots e_m\}$ is a collection of hyperedges, which are subsets of   $V$. The structure of a hypergraph is encoded in the $|V|\times|E|$ incidence matrix $\mathbf{H}$, where  $\mathbf{H}(v_i,e_j)=1$ if $v_i \in e_j$ and $0$ otherwise, for $i=1,2,\ldots,n$, $j=1,2,\ldots,m$. In probabilistic terms,  $\mathbf{H}(v_i,e_j)\in [0,1]$, i.e., its entry take probability value, reflecting the likelihood that a vertex belongs to a hyperedge.
A weighted hypergraph $G=(V, E, w)$, with each hyperedge $e\in E$ assigned a positive weight $w(e)$, $\mathbf{W}$ denoting the diagonal matrix of these weights. The degree of a hyperedge $e\in E$ is defined as  $\delta(e)=\sum_{v \in V}\mathbf{H}(v, e)$, while the degree of a vertex $v\in V$ is defined as $d(v)=\sum_{e \in E}w(e)\mathbf{H}(v, e)$. $\mathbf{D}_e$ and $\mathbf{D}_v$ represent the diagonal matrices composed of hyperedge degrees and vertex degrees, respectively.

In this paper, we generate a hypergraph for each view by sparse representation learning. Specifically, each vertex is depicted as a centroid, represented by a linear combination of other vertices with sparse coefficients. Vertices with non-zero coefficients to the centroid form a hyperedge, ultimately generating the hypergraph. Initially, we derive a similarity matrix for each view by solving a sparse representation problem to induce the hypergraph incidence matrix \cite{8662703}. Critically, the similarity matrix $\mathbf{A}$ is learned such that it inversely correlates with the distance between data points. To achieve this, the sparse representation problem for each view is formulated as follows:
\begin{align}
\label{a}
\min _{\mathbf{A}}  \sum_{i, j=1}^{n}\left\|\mathbf{x}_{i}-\mathbf{x}_{j}\right\|_{2}^{2} a_{i j}+\alpha \sum_{i=1}^{n}\left\|\mathbf{a}_{i}\right\|_{2}^{2}\notag\\
\text { s.t. } a_{i i}=0, a_{i j} \geq 0, \mathbf{1}^{T} \mathbf{a}_{i}=1,
\end{align}
where $\mathbf{a}_{i}^{T}$ is a row vector of $\mathbf{A}$, $i=1,\ldots,n$. The first term ensures that the larger distance $\left\|\mathbf{x}_{i}-\mathbf{x}_{j}\right\|_{2}^{2} $ between data points $x_i$ and $x_j$ corresponds to a smaller value $a_{i j}$.
The second term is the $l_2$-norm regularization term to assume the smoothness of the vectors $\mathbf{a}_{i}$.
The normalization $\mathbf{1}^{T} \mathbf{a}_{i}=1$ is equivalent to the sparse constraint on $\mathbf{A}$.

To solve problem   (\ref{a}), we define $c_{i j}=\left\|\mathbf{x}_{i}-\mathbf{x}_{j}\right\|_{2}^{2}$ and denote $\mathbf{c}_{i}$ as a vector with the $j$-th entry as $c_{i j}$. Given that the row vector $\mathbf{a}_{i}^{*}$ of the optimal solution $\mathbf{A}^{*}$ is sparse, it is assumed to have exactly $\sigma$ nonzero values. According to \cite{10.5555/3016100.3016174}, the optimal solution can be expressed as:
\begin{equation}
\label{eso}
a_{i j}^*=\left\{\begin{array}{cc}
\frac{c_{i, \sigma+1}-c_{i j}}{\sigma c_{i, \sigma+1}-\sum_{h=1}^{\sigma} c_{i h}}, & \text{if}\,\, j \leq \sigma, \vspace{2mm}\\
0, & \text{otherwise} .
\end{array}\right.
\end{equation}
In this paper, we consider each data point $v_i$ as a centroid vertex to generate hyperedge $e_i$. In addition to the centroid vertex, each hyperedge also includes $\sigma$ neighbors. Inspired by the probability incidence relationship \cite{2019Robust}, we establish the incidence relations by using $\sigma$ nonzero elements  from the $i$-th raw of the similarity matrix $\mathbf{A}^{*}$. Thereafter, we generate a hypergraph based on the set of hyperedges, with the incidence matrix $\mathbf{H}$  defined as follows:
\begin{equation}
\label{hyper}
\mathbf{H}(v_i,e_j)=\left\{\begin{array}{cc}
1, & \text{if}\,\,  v_i \in e_j, i=j,\vspace{1mm}\\
a_{ij}^*, & \text{if}\,\,  v_i \in e_j, i\neq j,\vspace{1mm}\\
0, & \text{otherwise}.
\end{array}\right.
\end{equation}
where $e_j$ is the hyperedge generated by $v_j$ and nonzero entries in the $j$th raw of $\mathbf{A}^{*}$. All hyperedges are considered to have equal weights here.  Consequently, a hypergraph  is fully characterized by the incidence matrix $\mathbf{H}$ and hyperedge weights matrix $\mathbf{W}$. Additionally, the initial hypergraph Laplacian matrix for each view is provided.

\begin{algorithm}
\caption{Hypergraph Generation}
\label{alg1}
\begin{algorithmic}[1]
\Statex \textbf{Input:} Feature matrix $\mathbf{X}$, constant $\sigma$
\Statex \textbf{Output:} Incidence matrix $\mathbf{H}$
\State Obtain $\mathbf{A}^*$ by   (\ref{eso})
\State Initialize $\mathbf{H}$ with zero of size $n \times n$
\For{$i = 1$ to $n$}
    \State Construct hyperedge $e_i$ with the centroid vertex $v_i$ and its $\sigma$ neighbors
    \State Update $\mathbf{H}$ for $e_i$ by (\ref{hyper})
\EndFor
\State \textbf{return} $\mathbf{H}$
\end{algorithmic}
\end{algorithm}

Generally, the dimensions of the incidence matrix are represented as $n\times m$, where $n$ represents the number of vertices and $m$ indicates the number of hyperedges. In our study, we first solve the sparse representation problem   (\ref{a}) for a dataset of size $n$ to obtain the similarity matrix $\mathbf{A}^*$ by (\ref{eso}) whose dimensions are $n\times n$. By regarding a data point as a centroid vertex and constructing a hyperedge $e$ by a centroid vertex and its neighbors, we generate the hypergraph with the incidence matrix whose dimensions are $n\times n$ by   (\ref{hyper}). As a result, the number of hyperedges is equal to the number of vertices, i.e., $m=n$. The hypergraph generation process is summarized in Algorithm \ref{alg1}. The computational complexity is $O(n^2)$ since $\sigma$ is a constant much smaller than $n$, meaning that the execution time is determined by the size of dataset.

\subsection{Hypergraph Spectral Clustering (HSC)}
\label{3.2}
Similar to simple graphs, 
we utilize Zhou's normalized Laplacian \cite{2006Learning}, which has been specifically adapted for $k$-way partitioning, and apply it in hypergraph partitioning as outlined in \cite{2006Learning}. Let $\left(V_{1}, \cdots, V_{k}\right)$ denote a k-way partition of $V$ such that $V_{1} \cup V_{2} \cup \cdots \cup V_{k}=V$, $V_{i} \cap V_{j}=\emptyset$ for all $i \neq j$. The boundary of $V_{i}$ in the hypergraph is denoted as $\partial V_{i}:=\left\{e \in E \mid e \cap V_{i} \neq \emptyset, e \cap V_{j} \neq \emptyset (j=1,\ldots,k,j\neq i)\right\}$, which consists of hyperedges cut. The sum of the degrees of vertices in $V_{i}$ is defined as the $volume$ $\operatorname{vol} V_{i}:=\sum_{v \in V_{i}} d(v)$. Additionally, the $volume$ of $\partial V_{i}$ is $\operatorname{vol} \partial V_{i}:=\sum_{e \in \partial V_{i}} w(e) \frac{|e \cap V_{i}|\left|e \cap V_{j}\right|}{\delta(e)}, j=1,\ldots,k,j\neq i$. The cost function considers the sum of the ratios of hyperedge cuts in each partition, which can be formulated as follows:
\begin{equation*}
c\left(V_{1}, \cdots, V_{k}\right)= \frac{\operatorname{vol} \partial V_{1}}{\operatorname{vol} V_{1}}+\frac{\operatorname{vol} \partial V_{2}}{\operatorname{vol} V_{2}}+\cdots+\frac{\operatorname{vol} \partial V_{k}}{\operatorname{vol} V_{k}}.
\end{equation*}
For a hypergraph, the normalized form of the hypergraph Laplacian is defined as   $\mathbf{\Delta}=\mathbf{I}_k-\mathbf{D}_{v}^{-\frac{1}{2}} \mathbf{H} \mathbf{W} \mathbf{D}_{e}^{-1} \mathbf{H}^{T} \mathbf{D}_{v}^{-\frac{1}{2}}$, where $\mathbf{I}_k$ is the $k \times k$ identity matrix. $\mathbf{\Theta}=\mathbf{D}_{v}^{-\frac{1}{2}} \mathbf{H }\mathbf{W} \mathbf{D}_{e}^{-1} \mathbf{H}^{T} \mathbf{D}_{v}^{-\frac{1}{2}}$ is the variant of the hypergraph Laplacian. Then the relaxation of the normalized hypergraph cut problem can be obtained by minimizing
\begin{align*}
&\frac{1}{2}\sum_{i=1}^{k} \sum_{e \in E} \sum_{\{u, v\} \subseteq e} \frac{w(e)}{\delta(e)}\left(\frac{f_i(u)}{\sqrt{d(u)}}-\frac{f_i(v)}{\sqrt{d(v)}}\right)^{2}\notag\\
&=\sum_{i=1}^{k} \mathbf{f}_i^{T}(\mathbf{I}_k-\mathbf{\Theta}) \mathbf{f}_i \notag\\
&=\operatorname{tr}\left(\mathbf{F}^{T} \mathbf{\Delta} \mathbf{F}\right),
\end{align*}
where $\operatorname{tr}(\cdot)$ denotes the trace of the matrix, and $\mathbf{F}=\left[\mathbf{f}_{1} \cdots \mathbf{f}_{k}\right] $ satisfies $\mathbf{F}^{T} \mathbf{F}=\mathbf{I}_k$. This condition enforces the orthogonality constraint between the multiple eigenvectors associated with $k$ smallest eigenvalues of $\mathbf{\Delta}$.
From \cite{2006Learning}, the original combinatorial optimization problem is relaxed into:
\begin{align}
\label{eq3}
& \min \limits_{\mathbf{F} \in \mathbb{R}^{n\times k } } \operatorname{tr}\left(\mathbf{F}^T \mathbf{\Delta} \mathbf{F}\right) \notag\\
& \text { s.t. } \quad \mathbf{F}^T \mathbf{F}=\mathbf{I}_k.
\end{align}
Since the sum of $\mathbf{\Delta}$ and $\mathbf{\Theta}$ is the identity matrix, (\ref{eq3}) is also equivalent to the following problem:
\begin{align}
\label{eq4}
& \max \limits_{\mathbf{F} \in \mathbb{R}^{n \times k}} \operatorname{tr}\left(\mathbf{F}^T \mathbf{\Theta} \mathbf{F}\right) \notag\\
& \text { s.t. } \mathbf{F}^T \mathbf{F}=\mathbf{I}_k.
\end{align}
The solution of   (\ref{eq4}) constitutes the orthogonal basis of the linear space spanned by the eigenvectors corresponding to $k$ largest eigenvalues of $\mathbf{\Theta}$. To enhance comprehension of hypergraph spectral clustering, we outline the computational steps as follows:
\begin{itemize}
    \item[i)] Generate the incidence matrix $\mathbf{H}$ and weight matrix $\mathbf{W}$.
    \item[ii)] Compute the degree matrices $\mathbf{D}_e$, $\mathbf{D}_v$, and then compute the Laplacian variant matrix
$\mathbf{\Theta}=\mathbf{D}_{v}^{-\frac{1}{2}} \mathbf{H} \mathbf{W} \mathbf{D}_{e}^{-1} \mathbf{H}^{T} \mathbf{D}_{v}^{-\frac{1}{2}}$.
    \item[iii)] Determine the eigenvectors associated with $k$ largest eigenvalues of $\mathbf{\Theta}$
and assemble them into $\mathbf{F} \in \mathbb{R}^{n \times k}$.
    \item[iv)] Normalize each row vectors of $\mathbf{F}$ and subsequently apply the standard $k$-means to these normalized vectors.
\end{itemize}

\subsection{Multi-view Hypergraph Spectral Clustering (MHSC)}

Based on the generated hypergraph, we propose an optimization function for multi-view hypergraph spectral clustering, which simultaneously considers hypergraph spectral clustering for each view and the consistency constraints between different views:
\begin{align}
\label{eq:mh}
\max _{\mathbf{F}^{(l)},\mathbf{F}^{*} \in \mathbb{R}^{n \times k}} & \sum_{l=1}^{r} \operatorname{tr}\left(\mathbf{F}^{{(l)}^T} \mathbf{\Theta}^{(l)} \mathbf{F}^{(l)}\right)-\sum_{l=1}^{r} \lambda_{l} D(\mathbf{F}^{(l)},\mathbf{F}^{*}) \notag\\
\text { s.t. } & \quad \mathbf{F}^{(l)^{T}} \mathbf{F}^{(l)}=\mathbf{I}_k,  \mathbf{F}^{*^{T}}\mathbf{F}^{*}=\mathbf{I}_k,
\end{align}
where $l=1,\ldots,r$.
The first term minimizes the ratio of hyperedge cuts in each partition by utilizing the eigenvectors associated with the $k$ largest eigenvalues of the hypergraph Laplacian, as discussed in Section \ref{3.2}. The second term enforces consistency by minimizing the discrepancy between similarity matrices for different views.
Similar to the similarity matrix defined for a simple graph \cite{NIPS2011_31839b03}, we use a linear kernel $\mathbf{K}_{\mathbf{F}^{(l)}}=\mathbf{F}^{(l)} \mathbf{F}^{(l)^{T}}$as the similarity matrix, where $\mathbf{F}^{(l)}$ is the eigenvector matrix associated with the hypergraph Laplacian generated from the $l$-th view. By disregarding constant additive and scaling factors, the pairwise discrepancy between two views can be quantified using the following measure:
\begin{align}
D\left(\mathbf{F}^{(l)}, \mathbf{F}^{(w)}\right)&=\left\|\frac{\mathbf{K}_{\mathbf{F}^{(l)}}}{\left\|\mathbf{K}_{\mathbf{F}^{(l)}}\right\|_{F}^{2}}-\frac{\mathbf{K}_{\mathbf{F}^{(w)}}}
{\left\|\mathbf{K}_{\mathbf{F}^{(w)}}\right\|_{F}^{2}}\right\|_{F}^{2}\notag\\
&=-\operatorname{tr}\left(\mathbf{F}^{(l)} \mathbf{F}^{(l)^{T}} \mathbf{F}^{(w)} \mathbf{F}^{(w)^{T}}\right).
\end{align}
Let $\mathbf{F}^{*}$ be the consensus view and $D\left(\mathbf{F}^{(l)}, \mathbf{F}^{*}\right)$ represent the  similarity discrepancy between the $l$-th view and the consensus view. To guide each view towards alignment with the consensus view, we aim to minimize $D\left(\mathbf{F}^{(l)}, \mathbf{F}^{*}\right)$ as much as possible. $\mathbf{F}^{*}$ also adheres to the orthogonality constraint. The second term is weighted by a trade-off parameter $\lambda_l$.

Problem   (\ref{eq:mh}) can be decomposed  into multiple subproblems by adopting an alternating iterative approach. Within each subproblem, the variables  $\mathbf{F}^{(l)}$ or $\mathbf{F}^{*}$ are contingent upon the eigenspace of the hypergraph Laplacian with mutual orthogonality constraints. The optimal consensus $\mathbf{F}^{*}$ is the unified eigenvector matrix we aim to obtain for multi-view clustering.
\subsection{MHSC Transformation on the Grassmannian Manifold}

To effectively tackle the independent subproblems of   (\ref{eq:mh}) for the variables $\mathbf{F}^{(l)}$ and $\mathbf{F}^{*}$, here are three key considerations: First, the variable of each subproblem is rooted in the eigenspace  spanned by the orthogonal eigenvectors of the view-specific hypergraph Laplacian, establishing a $k$-dimensional subspace representation. Notably, the collection of all such
$k$-dimensional subspaces constitutes the points on the Grassmannian manifold; Second, in Euclidean space, finding multiple eigenvectors $\mathbf{F}^{(l)}$ and $\mathbf{F}^{*}$ with orthogonality constraints  can result in approximation errors and convergence to local maxima \cite{NIPS2011_31839b03}.
In contrast, transforming the above constrained optimization to the Grassmannian manifold ensures that $\mathbf{F}^{(l)}$ and $\mathbf{F}^{*}$ in each iteration remain on the manifold, naturally maintaining the mutual orthogonality constraints and facilitating more efficient convergence to global maxima; Third, optimizing directly on the Grassmann manifold can be viewed as a form of dimensionality reduction compared to optimization in Euclidean space, as the variables on the manifold represent the $k$-dimensional subspaces (where $k<n$). The information of original problem is fully preserved, even with the reduced dimensionality. As a result, we devise an alternating iteration Riemannian optimization algorithm on the Grassmannian manifold to improve the efficiency and effectiveness for solving problem   (\ref{eq:mh}).

Concretely, a Grassmannian manifold $\mathcal{G}(k,n)$ is defined as the set of all linear $k$-dimensional subspaces in $\mathbb{R}^{n}$, where $k,n\in Z^{+}, k\leq n$. In its matrix representation, each point on $\mathcal{G}(k,n)$ is represented by the column space of a full rank matrix $\mathbf{X }\in \mathbb{R}^{n \times k}$. Each point on $\mathcal{G}(k,n)$ can be represented as follows:
$$
\mathcal{X}:= \operatorname{span}_k(\mathbf{X}),
$$
where $\operatorname{span}_k(\mathbf{X})$ represents the linear subspace spanned by the columns of $\mathbf{X}$. The Grassmannian manifold is defined as follows:
$$
\mathcal{G}(k, n)=\left\{\operatorname{span}_k(\mathbf{X}): \mathbf{X} \in \mathbb{R}^{n \times k}\right\}.
$$
If the orthogonality constraint is imposed on $\mathbf{X}$, it must satisfy $\mathbf{X}^{T} \mathbf{X}=\mathbf{I}_k$, where $\mathbf{I}_k$ denotes the $k \times k$ identity matrix. It ensures the following equivalence relation:
$$
\mathcal{X}:= \{\mathbf{X}\mathbf{Q},\mathbf{Q} \in \mathcal{O}(k)\},
$$
where $\mathcal{O}(k)$ is the $k$-order orthogonal group  that $\mathcal{O}(k)=\left\{{\mathbf{Q}} \in \mathbb{R}^{k \times k}|{\mathbf{Q}}^{T} {\mathbf{Q}}={\mathbf{I}_k}\right\}$. Under this equivalence relation, each point on $\mathcal{G}(k,n)$ forms an equivalence set. This allows us to treat the Grassmannian manifold as the quotient space of the Stiefel manifold $\mathcal{S}(k, n)$ \cite{Mishra}:
$$
\mathcal{G}(k, n):= S(k, n) / \mathcal{O}(k),
$$
where $
\mathcal{S}(k, n)=\left\{{\mathbf{X}} \in \mathbb{R}^{n \times k} \mid {\mathbf{X}}^{T} {\mathbf{X}}={\mathbf{I}_k}\right\}
$ is the set of matrices with $n \times k$ dimensions and orthonormal columns.

The constraint conditions in problem (\ref{eq:mh}) encompass matrices $\{
\mathbf{F}^{(l)} |\mathbf{F}^{(l)^{T}} \mathbf{F}^{(l)}=\mathbf{I}_k, l=1, \ldots, r\}$ and $\{
\mathbf{F}^{*} | \mathbf{F}^{*^{T}} \mathbf{F}^{*}=\mathbf{I}_k \}$, with their column vectors are mutually orthogonal, which make it challenging to discover closed-form solutions. To overcome this bottleneck, various relaxation methods have been proposed \cite{6805155,7451227}. The solution is  approximated by leveraging the explicit eigenvectors associated with the $k$ largest  eigenvalues of $ \mathbf{F}^{(l)} $ and  $ \mathbf{F}^{*}$. As previously observed, the $k$-dimensional linear subspaces of the eigenspace, spanned by these
mutually orthogonal eigenvectors, inherently constitute the Grassmannian manifold. The intrinsic  geometry of this manifold ensures that the points in each iteration consistently reside on the manifold. To facilitate further discussion, we rewrite the objective function of problem   (\ref{eq:mh}) 
as $f(\mathbf{F}^{(1)}, \ldots, \mathbf{F}^{(r)}, \mathbf{F}^{*})$ and reach the following lemma.

\vspace{2mm}
\begin{lemma}\label{le}
The objective function of problem  (\ref{eq:mh})  is orthogonal transformation invariant, i.e., for any $\mathbf{Q} \in \mathcal{O}(k)$ we have
\begin{equation}
f(\mathbf{F}^{(1)}\mathbf{Q}, \ldots, \mathbf{F}^{(r)}\mathbf{Q}, \mathbf{F}^{*}\mathbf{Q})=f(\mathbf{F}^{(1)}, \ldots, \mathbf{F}^{(r)}, \mathbf{F}^{*}).
\end{equation}
\end{lemma}

\textbf{\emph{Proof}}:
 Let $f_1(\mathbf{F}^{(l)})$  represent $\operatorname{tr}\left(\mathbf{F}^{{(l)}^T} \mathbf{\Theta}^{(l)} \mathbf{F}^{(l)}\right)$ and $f_2(\mathbf{F}^{(l)},\mathbf{F}^*)$  represent $-\lambda_{l}D(\mathbf{F}^{(l)},\mathbf{F}^{*})$, and we have $f(\mathbf{F}^{(1)}, \ldots, \mathbf{F}^{(r)}, \mathbf{F}^{*})=\sum_{l=1}^{r}f_1(\mathbf{F}^{(l)})+\sum_{l=1}^{r}f_2(\mathbf{F}^{(l)},\mathbf{F}^*)$.
Utilizing the fact that the trace operator is invariant under
cyclic permutations and that $\mathbf{Q}$ is an orthogonal matrix, it follows that
\begin{equation*}
\begin{aligned}
f_1(\mathbf{F}^{(l)}\mathbf{Q})&=\operatorname{tr}\left(\mathbf{Q}^T\mathbf{F}^{(l)^{T}} \mathbf{\Theta}^{(l)} \mathbf{F}^{(l)}\mathbf{Q}\right)\\
&=\operatorname{tr}\left(\mathbf{Q}^{-1}\mathbf{F}^{(l)^{T}} \mathbf{\Theta}^{(l)} \mathbf{F}^{(l)}\mathbf{Q}\right)\\
&=f_1(\mathbf{F}^{(l)})
\end{aligned}
\end{equation*}
and
\begin{equation*}
\begin{aligned}
f_2(\mathbf{F}^{(l)}\mathbf{Q},\mathbf{F}^*\mathbf{Q})&=\lambda_{l}\operatorname{tr}\left(\mathbf{F}^{(l)}\mathbf{Q} \mathbf{Q}^T \mathbf{F}^{(l)^{T}} \mathbf{F}^{*} \mathbf{Q} \mathbf{Q}^T \mathbf{F}^{*^{T}}\right)\\
&=f_2(\mathbf{F}^{(l)},\mathbf{F}^*),
\end{aligned}
\end{equation*}
where $l=1,\ldots,r$. It is evident that one can confirm the equivalence: $f(\mathbf{F}^{(1)}\mathbf{Q}, \ldots, \mathbf{F}^{(r)}\mathbf{Q}, \mathbf{F}^{*}\mathbf{Q})=f(\mathbf{F}^{(1)}, \ldots, \mathbf{F}^{(r)}, \mathbf{F}^{*})$.
$\hfill\square$
\vspace{2mm}

\begin{remark}
The set of matrices adhering to the orthogonality constraint constitutes the Stiefel manifold $\mathcal{S}(k, n)$. Nonetheless, as indicated in Lemma \ref{le}, the optimization of problem   (\ref{eq:mh}) confined exclusively to the Stiefel manifold may yield non-unique solutions, stemming from the equal objective function value across the equivalent classes.  In contrast, on the Grassmannian manifold $\mathcal{G}(k,n)$, a set of equivalence classes are represented as a point, ensuring the uniqueness of the solution.
\end{remark}

\vspace{2mm}
\begin{remark}
Let $[\mathbf{F}^{(l)}]\in \mathcal{G}(k,n)$ and $[\mathbf{F}^{*}]\in \mathcal{G}(k,n)$ denote the set of equivalence classes of $\mathbf{F}^{(l)}$ and $\mathbf{F}^{(l)}$, respectively. According to Lemma \ref{le}, the values of $f_1(\mathbf{F}^{(l)})$ and $D(\mathbf{F}^{(l)},\mathbf{F}^{*})$ still remain the same across the equivalence classes $[\mathbf{F}^{(l)}]$ and $[\mathbf{F}^{*}]$.
For simplicity, we still refer to these equivalence classes using the original notations $\mathbf{F}^{(l)}$ and $\mathbf{F}^{*}$.
\end{remark}
\vspace{2mm}

Finally, we transform the optimization problem (\ref{eq:mh}) with orthogonality constraints into  an unconstrained optimization problem  on the Grassmannian manifold:

\begin{equation}
\label{eq:mo}
\begin{aligned}
\max_{ \mathbf{F}^{(l)}, \mathbf{F}^{*} \in \mathcal{G}(k,n)}
&  \sum_{l=1}^{r} \operatorname{tr}\left(\mathbf{F}^{{(l)}^T} \mathbf{\Theta}^{(l)} \mathbf{F}^{(l)}\right)-\sum_{l=1}^{r} \lambda_{l} D(\mathbf{F}^{(l)},\mathbf{F}^{*}).
\end{aligned}
\end{equation}

\section{Optimization Algorithm}\label{s3}

In this section, we elucidate the process of solving problem   (\ref{eq:mo}) through an alternating iterative Riemann optimization strategy, due to the coupling of all variables. This problem is  decomposed into several subproblems, each concentrating on a single variable while keeping the remaining variables fixed. We proceed to iteratively update the variable within each subproblem.

Problem  (\ref{eq:mo}) is situated in subspaces endowed
with  Riemannian geometric structure. To tackle such  manifold optimization problem, various algorithms have been crafted.  Notably, the work of Edelman et al. \cite{1998The} has laid the foundation for the Riemannian geometry on the Stiefel manifold $S(k, n)$ and Grassmannian manifold $\mathcal{G}(k,n)$, including the Riemannian gradient and Hessian in these subspaces. Moreover, the Riemannian gradient can be derived from the Euclidean counterpart by the following lemma.

\vspace{2mm}
\lemmawithcite{1998The}
\label{lemma}
Let $\mathcal{M}$ be a Riemannian submanifold of a Euclidean space E. Suppose $\bar{f}: E \rightarrow \mathbb{R}$ is a function with Euclidean gradient $\operatorname{grad} \bar{f}(x)$ at point $x \in \mathcal{M}$. Then the Riemannian gradient of $f:=\left.\bar{f}\right|_{\mathcal{M}}$ is given by $\operatorname{grad} f(x)=\mathrm{P}_{x} \operatorname{grad} \bar{f}(x)$, where $\mathrm{P}_{x}$ denotes the orthogonal projection onto the tangent space $T_{x} \mathcal{M}$.
\vspace{2mm}

Following Lemma \ref{lemma}, we  determine the  Riemannian gradients, denoted as
$\operatorname{grad}_{\mathbf{F}^{(l)}} f$ and $\operatorname{grad}_{\mathbf{F}^{*}}f$,  on the Grassmannian manifold  for each variable $\mathbf{F}^{(l)}$ and $\mathbf{F}^{*} \in \mathcal{G}(k,n)$, $l=1, \ldots, r$. Armed with the objective function and its gradients, we proceed to apply the Riemannian Trust-Region (RTR) solver to each subproblem, leveraging  a well-established ManOpt toolbox\footnote{{https://www.manopt.org/}}.

\subsection{Update of $\mathbf{F}^{(l)} (l=1,\ldots,r)$ with Fixing Others}

Given the consensus view $\mathbf{F}^{*}$ and other views $\mathbf{F}^{(w)}(w\neq l)$, the subproblem for view $l$ is solved as follows:
\begin{equation}
\label{eq:12}
\max _{\mathbf{F}^{(l)} \in \mathcal{G}(k,n)} \operatorname{tr}\left(\mathbf{F}^{(l)^{T}} \mathbf{\Theta}^{(l)} \mathbf{F}^{(l)}\right)+\lambda_{l} \operatorname{tr}\left(\mathbf{F}^{(l)} \mathbf{F}^{(l)^{T}} \mathbf{F}^{*} \mathbf{F}^{*^{T}}\right).
\end{equation}

It is evident that problem   (\ref{eq:12}) is contingent on the $k$-dimensional subspaces of the eigenspace, which is spanned by the mutually orthogonal eigenvectors  of
${\Theta}^{(l)}+\lambda_{l}\mathbf{F}^{*} \mathbf{F}^{*^{T}}$.  We employ RTR on $\mathcal{G}(k,n)$ to derive a solution $\mathbf{F}^{(l)}$,  where its columns provide an orthogonal basis for the dominant subspace. Additionally, the Riemannian gradient of the objective function   (\ref{eq:12}) is given by
\begin{equation}
\operatorname{grad}_{\mathbf{F}^{(l)}}f =2\left(\mathbf{I}-\mathbf{F}^{(l)} \mathbf{F}^{(l)^{T}}\right)
\tilde{\mathbf{\Theta}}^{(l)}
\mathbf{F}^{(l)}
\end{equation}
with $\tilde{\mathbf{\Theta}}^{(l)}= \left(\Theta^{(l)}+\lambda_{l}\mathbf{F}^{*} \mathbf{F}^{*^{T}}\right)$.

\begin{figure*}[htbp]
  \centering
  \includegraphics[width=0.88\linewidth]{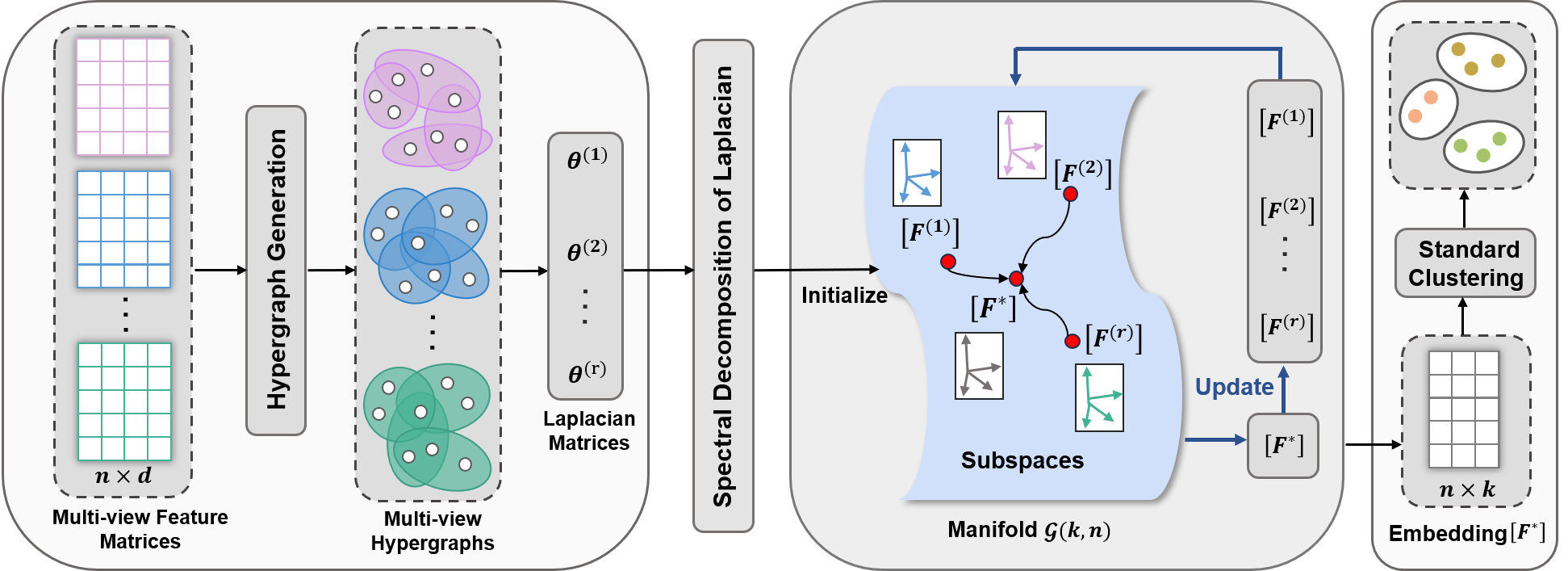}\\
  \caption{Schematic diagram of the multi-view hypergraph spectral clustering on the Grassmannian manifold. $\mathbf{\Theta}^{(l)}$ denotes the hypergraph laplacian variant matrix of $l$-th view, and the initialized eigenvector matrix $\mathbf{F}^{(l)}$ can be obtained through its spectral decomposition,
  where $l=1,\ldots,r$. $[\mathbf{F}^{(l)}]$ denote the equivalence classes of
  $\mathbf{F}^{(l)}$. Each point on manifold $\mathcal{G}(k,n)$ corresponds to a set of  k-dimensional dominant subspaces of the eigenspace.
}
  \label{fig1}
\end{figure*}
\subsection{Update of $\mathbf{F}^{*}$ with Fixing Others}
Given $\mathbf{F}^{(1)},\mathbf{F}^{(2)},\ldots, \mathbf{F}^{(r)}$, the subproblem for the consensus view $\mathbf{F}^{*}$ is solved as follows:
\begin{equation}
\label{eq:14}
\max _{\mathbf{F}^{*} \in \mathcal{G}(k,n)} \sum_{l=1}^r \lambda_{l} \operatorname{tr}\left(\mathbf{F}^{(l)} \mathbf{F}^{(l)^{T}} \mathbf{F}^{*} \mathbf{F}^{*^{T}}\right).
\end{equation}
Similarly, problem   (\ref{eq:14}) hinges on the $k$-dimensional subspaces of eigenspace, which are spanned by the mutually orthogonal eigenvectors of $\sum_{l=1}^{r}\lambda_{l} \mathbf{F}^{(l)} \mathbf{F}^{(l)^{T}}$.  We apply RTR on $\mathcal{G}(k,n)$ to derive a solution $\mathbf{F}^{*}$,  where its columns constitute an orthogonal basis for the dominant subspace. Moreover, the Riemannian gradient of the objective function   (\ref{eq:14}) is given by
\begin{equation}
\operatorname{grad}_{\mathbf{F}^{*}}f =2\left(\mathbf{I}-\mathbf{F}^{*} \mathbf{F}^{*^{T}}\right)
\mathbf{\Theta}^{*} \mathbf{F}^{*}
\end{equation}
with $\mathbf{\Theta}^{*}=\sum_{l=1}^{r}\lambda_{l} \mathbf{F}^{(l)} \mathbf{F}^{(l)^{T}}$.

\subsection{Adaptive Update of the Parameter $\lambda_{l}$}

According to \cite{8662703}, the optimal solution of problem   (\ref{eq:mo}) ensures that the number of connected components in the hypergraph exactly matches the number of clusters, $k$. Let $val=\operatorname{tr}(\mathbf{F}^{*^{T}} \mathbf{\Delta}^* \mathbf{F}^*)$, which equals $\sum_{i=1}^k \vartheta_i\left(\mathbf{\Delta}^*\right)$, where $\vartheta_i\left(\mathbf{\Delta}^*\right)$ denotes the $i$-th smallest eigenvalue of $\mathbf{\Delta}^*$.
Achieving $val=0$ indicates that the corresponding $\mathbf{F}^*$ is an ideal consensus,  leading to a direct partitioning of data points into $k$ clusters \cite{8662703}.
In this scenario, we have $\mathbf{\Delta}^*=\mathbf{I}-\mathbf{\Theta}^{*}$ with $\mathbf{\Theta}^{*}=\sum_{l=1}^{r}\lambda_{l} \mathbf{F}^{(l)} \mathbf{F}^{(l)^{T}}$. Besides the specific structure of $\mathbf{F}^{(l)}$ and the sizes of $r$ and $k$, the adjustment of $\lambda_l$ is a key factor influencing  the eigenvalues of $\mathbf{\Delta}^*$. By adaptively updating the parameter  $\lambda_l$ in each iteration, the goal of $val=0$ can be achieved. For this purpose, given the initial value of $\lambda_l$, there is no need for additional tuning; an automatic conditional update mechanism is introduced. After each iteration, if the condition $val=0$ is satisfied, it implies that the current configuration captures the ideal clustering structure, eliminating the need to adjust $\lambda_{l}$. Conversely, if the condition is not satisfied, indicating a mismatch with the desired number of clusters, $\lambda_{l}$ is then adaptively reduced to $\mathbf{\lambda}_l/2$.
\begin{algorithm}
\caption{Optimization Algorithm for MHSCG}
	\label{alg2}
	\begin{algorithmic}[1]
		\Require Multi-view feature matrices $\mathbf{X}^{(l)}$ ($l=1,\ldots,r$), cluster number $k$
		\Ensure  Cluster assignments $C_1,\ldots,C_k$
        \State Generate  $\mathbf{H}$ by using Eq.   (\ref{hyper}) for each view.
        \State Initialize: maximal iteration $T$, $\lambda_l={1} (l=1,\ldots,r)$, $\epsilon=10^{-5}$, view-specific Laplacian $\mathbf{\Theta}^{(l)}$ and $\mathbf{F}^{(l)}$, consensus Laplacian $\mathbf{\Theta}^{*}$ and $\mathbf{F}^{*}$
        \For{each iteration $t\in [1,T]$}
		\For {each view $l\in [1,r]$}
		\State Fix $\mathbf{F}_t^*$,

\quad Update $\mathbf{F}_{t+1}^{(l)}$ by applying RTR
on
  (\ref{eq:12})
		\EndFor
        \State Fix $\mathbf{F}_{t+1}^{(1)},\ldots,\mathbf{F}_{t+1}^{(r)}$,

\hspace*{-3mm} Update $\mathbf{F}_{t+1}^{*}$ by applying RTR
on
  (\ref{eq:14})
        \State Compute $val=\operatorname{tr}(\mathbf{F}_{t+1}^{*^{T}} \mathbf{\Delta}_{t+1}^* \mathbf{F}_{t+1}^*) $
        \While{$|val|>\epsilon$}
        \State $ \mathbf{\lambda}_l \gets \mathbf{\lambda}_l/ 2$
        \State $\mathbf{F}_{t+1}^* \gets \mathbf{F}_{t}^*$
        \EndWhile
        \EndFor
        \State Normalize each row of $\mathbf{F}^*$ to get $\mathbf{F}_{norm}^*$
        \State Run $k$-means on each row of $\mathbf{F}_{norm}^*$ into $C_1,\ldots,C_k$
	\end{algorithmic}
\end{algorithm}

The optimization problem   (\ref{eq:mo}) corresponds to  multi-view hypergraph spectral clustering on the Grassmannian manifold (MHSCG). And the schematic framework of the proposed method for MHSCG is depicted in Fig. \ref{fig1}.
The entire optimization procedure for solving problem   (\ref{eq:mo})
is shown in Algorithm \ref{alg2}. In model   (\ref{eq:mo}), the parameter $\lambda_{l}$ is crucial for balancing the consistency of each individual view $\mathbf{F}^{(l)}$ with the consensus view $\mathbf{F}^{*}$, while also preserving the specific information inherent to each view. A larger value of $\lambda_{l}$ tends to align the representation of each view more closely with the consensus view. However, large $\lambda_{l}$ might neglect the distinct valuable insights from each view.
Therefore, the impact of the $\lambda_{l}$ value on clustering performance necessitates consideration of the specific characteristics and complexity of the multi-view data. For small $\lambda_{l}$, the algorithm may be prone to getting stuck in local optima, resulting in slower convergence and longer running times. As $\lambda_{l}$ increases, the algorithm is more likely to find the global optimum, speeding up convergence and reducing running times. When $\lambda_{l}$ is too large, it may overly emphasize the clustering consistency constraint regularization term, resulting in a slight increase in running time further.
The impact of different initial $\lambda_{l}$ values on clustering results is discussed at the end of Section \ref{s5.3}.

\subsection{Convergence Analysis}
\label{Convergence}
As outlined in Algorithm \ref{alg2}, we develop a novel algorithm that combines Riemannian optimization with an alternating iterative technique to solve problem   (\ref{eq:mo}).
All variables $\mathbf{F}^{(1)}, \cdots, \mathbf{F}^{(r)}, \mathbf{F}^{*} \in \mathcal{G}(k,n)$, meaning they belong to the same set of equivalence classes consisting of matrices with $n \times k$  dimensions and orthonormal columns on the Grassmannian manifold.
For a matrix $\mathbf{X}\in \mathbb{R}^{n \times k}$ that satisfies $\mathbf{X}^{T} \mathbf{X}=\mathbf{I}_k$, we endow it with the Frobenius norm:
$$\|\mathbf{X}\|_F=
\sqrt{\operatorname{tr}\left(\mathbf{X}^{T} \mathbf{X}\right)}=\sqrt{\operatorname{tr}\left(\mathbf{I}_k\right)}=\sqrt{k}.$$
Exploiting the properties of matrix norms, we observe that the set comprising all column orthogonal matrices is both bounded and closed. According to Henie-Borel Theorem, in a finite dimensional topological space, a set being compact is equivalent to being bounded and closed.  Consequently, the topological space in which all the variables reside is compact.  Moreover, the objective function of  problem (\ref{eq:mo}) is continuous, affirming the existence of the maximum or the minimum for it.

Let $f^t$ represent  the function value of problem   (\ref{eq:mo})
corresponding to the $t$-th iteration:

\begin{equation}
f^t=f(\mathbf{F}_t^{(1)},\cdots,\mathbf{F}_t^{(l)},\cdots,\mathbf{F}_t^{(r)},\mathbf{F}_t^*).
\end{equation}
Then, during the process from the $t$-th iteration to the ($t+1$)-th iteration, the variable is updated according to its corresponding subproblem.

\textbf{Update $\mathbf{F}_t^{(l)} (l=1,\ldots,r)$}.
Suppose that only updating $\mathbf{F}_t^{(l)}$, while keeping other variables fixed, it leads to a change in the objective function for problem (\ref{eq:mo}) by $a_t^{(l)}$:

\begin{align}
f(\mathbf{F}_{t+1}^{(1)},\mathbf{F}_{t}^{(2)},\cdots,\mathbf{F}_t^{(r)},\mathbf{F}_t^*)
&=f^t+a_t^{(1)}, \nonumber\\
f(\mathbf{F}_{t+1}^{(1)},\mathbf{F}_{t+1}^{(2)},\cdots,\mathbf{F}_t^{(r)},\mathbf{F}_t^*)
&=f^t+a_t^{(1)}+a_t^{(2)}, \nonumber\\
&\vdots \nonumber\\
f(\mathbf{F}_{t+1}^{(1)},\mathbf{F}_{t+1}^{(2)},\cdots,\mathbf{F}_{t+1}^{(r)},\mathbf{F}_t^*)
&=f^t+\sum_{l=1}^{r}a_t^{(l)}.
\end{align}
To obtain $\mathbf{F}_{t+1}^{(l)}$, we utilize RTR to solve the subproblem   (\ref{eq:12}), which involves approximating the objective function with a quadratic model in the vicinity of the tangent space of the variable $\mathbf{F}_t^{(l)}$.
The solution in the tangent space is then mapped back to the Grassmannian manifold by using a retraction operator, yielding $\mathbf{F}_{t+1}^{(l)}$.
In the tangent space, the trust-region method approximates the original optimization problem by solving a local quadratic programming subproblem to identify the steepest ascent direction.
Consequently, the change in objective function on the Grassmannian manifold must be nonnegative, i.e.,
$a_t^{(l)}\geq 0$ and $\sum_{l=1}^{r}a_t^{(l)}\geq 0$.

\textbf{Update $\mathbf{F}_t^{*}$}. After obtaining $\mathbf{F}_{t+1}^{(1)},\cdots,\mathbf{F}_{t+1}^{(l)},\cdots,\mathbf{F}_{t+1}^{(r)}$  at ($t+1$)-th iteration, suppose that only updating $\mathbf{F}_t^{*}$ while keeping other variables fixed leads to a change in the objective function for problem   (\ref{eq:mo}) by an amount denoted as $b_t$:
\begin{equation}
f^{t+1}=f(\mathbf{F}_{t+1}^{(1)},\cdots,\mathbf{F}_{t+1}^{(l)},\cdots,\mathbf{F}_{t+1}^{(r)},\mathbf{F}_{t}^*)+b_t.
\end{equation}
Analogously, when the variable  $\mathbf{F}_t^{*}$ is updated  by solving subproblem   (\ref{eq:14}) with RTR, it results in an increase or at least maintains the objective function value. Therefore, it can be inferred that $b_t\geq 0$.

Moreover, the variation in the objective function for problem   (\ref{eq:mo}) throughout the (${t+1}$)-th iteration  can be expressed as
\begin{equation}
f^{t+1}=f^t+\sum_{l=1}^{r}a_t^{(l)}+b_t.
\end{equation}
Consequently, for the variable sequence produced  by $T$ iterations, the objective function in problem   (\ref{eq:mo}) shows monotonic growth and is bounded by an upper limit. Given the compactness of the variable set, it can be deduced that the objective function value will converge towards its limiting value.

\section{Experimental Results}\label{s4}

To test the effectiveness of the proposed algorithm, we conduct experiments over four  multi-view datasets in comparison to seven baselines. These datasets include two text datasets: 3sources, BBC; a handwritten dataset: Mfeat Digits; an object recognition dataset: MSRCv1.
The evaluation metrics for clustering are: accuracy (ACC), normalized mutual information (NMI), F-score, and adjusted rand index (Adj-RI).
We present the experimental results and the corresponding low-dimensional visializations of embeddings.
Finally, we analyze the impact of initial parameter values and the convergence properties of the proposed algorithm, as well as an ablation study and statistical comparisons.

\subsection{Datasets}

We utilize four multi-view datasets. The descriptions of these datasets are summarized as follows: 


\textbf{3sources}\footnote{{http://mlg.ucd.ie/datasets/3sources.html}} is collected from $3$ online news sources: BBC, Reuters, and the Guardian. It contains a corpus of $169$ news stories, with $6$ thematic labels: business, entertainment, health, politics, sport, and technology.

\textbf{BBC}\footnote{{http://mlg.ucd.ie/datasets/segment.html}\label{BBC}} is collected from synthetic text collection with $4$ distinct views. It comprises a compilation of $685$ documents from the BBC news website. It encompasses $5$ categories: business, entertainment, politics, sport and tech.

\begin{table*}[htbp]
\vspace{-2mm}
\caption{Clustering Results of Different Methods on Multi-view Datasets}
\vspace{-2mm}
\centering
\begin{tabular}{cccccc}
\toprule
Dataset    & Method         & ACC(\%)         & NMI(\%)        & F-score(\%)    & Adj-RI(\%)     \\
\midrule
\multirow{8}*{3sources}   & HSC & 50.41$\pm$0.66 & 41.64$\pm$0.44 & 42.87$\pm$0.56 & 24.94$\pm$0.65 \\
           & CoRegSC        & 53.17$\pm$2.56 & 48.96$\pm$1.94 & 44.83$\pm$2.35 & 29.78$\pm$0.03 \\
           & WMSC           & 50.09$\pm$2.85 & 47.23$\pm$2.32 & 45.25$\pm$2.50 & 30.08$\pm$3.21 \\
           & GMC            & \underline{69.23}$\pm$0.00 & \underline{62.16}$\pm$0.00 & \underline{60.47}$\pm$0.00 & \underline{44.32}$\pm$0.00 \\
           & CGL  & 54.26$\pm$1.67 &51.41$\pm$1.25 &51.03$\pm$0.23 &37.40$\pm$1.37\\
           & MCLES & 62.90$\pm$2.12 &50.77 $\pm$2.84 &54.57$\pm$4.62 &34.47$\pm$6.04\\
           & CDMGC  & 43.43$\pm$0.04 &20.69$\pm$0.06 &36.42$\pm$0.00 &2.09$\pm$0.02\\
           & MHSCG           & \textbf{84.89}$\pm$6.01 & \textbf{74.76}$\pm$3.73 & \textbf{78.46}$\pm$5.64 & \textbf{72.60}$\pm$6.95 \\
\midrule
\multirow{8}*{BBC}        & HSC & 47.22$\pm$0.40 & 22.78$\pm$0.48  & 36.73$\pm$0.36 & 12.24$\pm$0.52  \\
           & CoRegSC        & 65.37$\pm$0.14 & 48.27$\pm$0.07 & 49.98$\pm$0.11 & 33.33$\pm$0.20 \\
           & WMSC           & 64.96$\pm$0.00 & 47.56$\pm$0.03 & 49.42$\pm$0.01 & 32.60$\pm$0.03 \\
           & GMC            & 69.34$\pm$0.00 & \underline{ 56.28}$\pm$0.00 & \underline{63.33}$\pm$0.00 & \underline{47.89}$\pm$0.00 \\
           & CGL  & 57.37$\pm$0.00 &40.66$\pm$0.00 &42.29$\pm$0.00 &19.90$\pm$0.00\\
           & MCLES  & \underline{70.85}$\pm$1.60 &56.03 $\pm$2.60 & 62.34$\pm$2.70 & 46.67$\pm$4.30\\
           & CDMGC  & 35.12$\pm$0.01 &5.88$\pm$0.01 &37.33$\pm$0.00 &0.06$\pm$0.00\\
           & MHSCG           & \textbf{90.71}$\pm$3.53 & \textbf{76.80}$\pm$1.61 & \textbf{84.70}$\pm$2.58 & \textbf{80.02}$\pm$3.46 \\
\midrule
\multirow{8}*{Mfeat Digits} & HSC & 93.90$\pm$0.00 & 87.75$\pm$0.00 & 88.34$\pm$0.00 & 87.05$\pm$0.00 \\
           & CoRegSC        & 81.20$\pm$0.03 & 83.20$\pm$0.04 & 78.36$\pm$0.04 & 75.88$\pm$0.04 \\
           & WMSC           & 85.10$\pm$0.01 & 84.27$\pm$0.02 & 80.35$\pm$0.02 & 78.17$\pm$0.02 \\
           & GMC            & 88.30$\pm$0.00 & 90.74$\pm$0.00 & 86.83$\pm$0.00 & 85.31$\pm$0.00 \\
           & CGL  &  \textbf{96.85}$\pm$0.00 &\textbf{93.02}$\pm$0.00 &\textbf{93.84}$\pm$0.00 &\textbf{93.15}$\pm$0.00\\
           & MCLES & 84.98$\pm$5.67 &83.29$\pm$0.96 &79.97$\pm$4.19 &77.65$\pm$4.77\\
           & CDMGC  & 85.55$\pm$0.00 &90.45$\pm$0.00 &85.07$\pm$0.00 &83.27$\pm$0.00\\
           & MHSCG           & \underline{{95.22}}$\pm$3.52 & \underline{{92.20}}$\pm$1.04 & \underline{{92.15}}$\pm$2.89 & \underline{{91.27}}$\pm$3.26 \\
\midrule
\multirow{8}*{MSRCv1}     & HSC & 70.05$\pm$0.14 & 68.09$\pm$0.27 & 62.52$\pm$0.13 & 55.76$\pm$0.15 \\
           & CoRegSC        & 84.76$\pm$0.00 & 77.02$\pm$0.00 & 75.48$\pm$0.00 & 71.44$\pm$0.00 \\
           & WMSC           & 76.33$\pm$0.34 & 71.89$\pm$0.04 & 68.77$\pm$0.07 & 63.56$\pm$0.09 \\
           & GMC            & 74.76$\pm$0.00 & 74.45$\pm$0.00 & 67.20$\pm$0.00 & 60.94$\pm$0.00 \\
           & CGL  & 85.71$\pm$0.00 &77.29$\pm$0.00 &75.26$\pm$0.00 &71.20$\pm$0.00\\
           & MCLES & \underline{87.32}$\pm$4.90 &\textbf{81.57} $\pm$1.93 &\underline{78.51}$\pm$3.16 &\underline{74.93}$\pm$3.79 \\
           & CDMGC  & 75.90$\pm$0.02 &71.67$\pm$0.01 &67.09$\pm$0.03 &61.18$\pm$0.03\\
           & MHSCG           & \textbf{89.52}$\pm$0.00 & \underline{79.75}$\pm$0.00 & \textbf{80.56}$\pm$0.00 & \textbf{77.42}$\pm$0.00\\
\bottomrule
\addlinespace[0.3mm]
\multicolumn{6}{l}{Bold indicates the best result and underline signifies the second best result.}\\
\end{tabular}
\label{tab2}
\end{table*}


\textbf{Mfeat Digits}\footnote{http://archive.ics.uci.edu/ml/datasets/Multiple+Features} dataset is a collection of handwritten digits with $6$ distinct views. It consists of $10$ categories ranging from 0 to 9. Total 2000 instances have been digitized into binary images.

\textbf{MSRCv1}\footnote{http://research.microsoft.com/en-us/project/} is an object dataset comprised of $210$ images, encompassing $7$ categories: tree, building, airplane, cow, face, car and bicycle. Each image is described by $5$ different views. 

\subsection{Compared Methods and Experinmetal Setup}
We compare our method with the following seven state-of-the-art methods.

\textbf{HSC} \cite{2006Learning}
serves as the foundational baseline. It applies conventional hypergraph spectral clustering to individual views, subsequently selecting the optimal clustering results.

\textbf{CoregSC} \cite{NIPS2011_31839b03} is an advanced multi-view spectral clustering method that ensures consensus among various views by incorporating co-regularization terms into its multi-view spectral clustering formulation.

\textbf{WMSC} \cite{Zong} leverages spectral perturbation in a weighted multi-view spectral clustering framework, utilizing the maximum gauge angle to measure spectral clustering performance variance among different views.

\textbf{GMC} \cite{8662703} employs a graph-based approach for multi-view clustering, automatically assigning weights to individual view graphs to construct an integrated fusion graph. It learns the specific graph for each view as well as the optimal fusion graph.

\textbf{CGL} \cite{8970909} offers a cohesive framework that adeptly leverages regularization to mitigate inconsistency among views while concurrently learning a unified graph. It takes into account both the consistency and the inconsistency information across various views.

\textbf{MCLES} \cite{2020MultiChen} is a method that concurrently learns the global structure and the cluster indicator matrix in a unified optimization framework.

\textbf{CDMGC} \cite{2021Measuring} introduces a unified framework to harness both the consistency and the inherent diversity of multi-view data in a synchronized manner.

We meticulously calibrate the parameters for all baselines, following the setting in their respective papers. To minimize the impact of randomness, clustering process of each algorithm was executed 30 times.
We record the mean and standard deviation of  clustering results. All experiments have been conducted 
with the same computational setup for all baselines.

\subsection{Clustering Results and Discussions}
\label{s5.3}
All clustering results from different algorithms are summarized in Table \ref{tab2}. It is worth noting that a higher value of each metric indicates the better clustering performance.

\begin{itemize}
    \item[i)]From Table \ref{tab2}, when compared to the hypergraph spectral clustering method, namely HSC, the other multi-view spectral clustering methods demonstrate enhanced clustering performance on the majority of datasets. This indicates the necessity of establishing a comprehensive multi-view clustering method for multi-view datasets, capable of integrating valuable information from different views of data, a capability that single-view methods cannot have.
    \item[ii)] In comparison to six multi-view clustering methods including CoRegSC, WMSC, GMC, CGL, MCLES, and CDMGC, the proposed method consistently delivers superior clustering results across the majority of datasets. This advantage is especially pronounced in the 3sources and BBC datasets, with improvements ranging from 15 to 40 percentage. 
    \item[iii)] There is one notable exception on the Mfeat Digits dataset, where the multi-view CGL method  achieves the highest clustering accuracy, slightly outperforming the proposed method, which secures the second-best performance. This is due to some views of Mfeat Digits are characterized by discrete features, while others are characterized by continuous features, leading to inconsistencies in data representation. The CGL method incorporates both inconsistency and consistency within its model design, whereas others primarily concentrate on maintaining consistency across views.
\end{itemize}

Overall, these observations demonstrate the superiority of the proposed method. We argue that there are three main reasons.
First,  the generated hypergraph for each view based on sparse representation learning captures potential high-order relationships and eliminates outliers and noise. Second, the proposed multi-view hypergraph spectral clustering optimization function addresses spectral clustering for each view and enforces consistency across different views simultaneously, improving the accuracy and robustness of clustering. Third, the hypergraph spectral clustering problem with orthogonal constraints is transformed into an unconstrained problem on the Grassmannian manifold. This transformation
reduces the variable dimensionality, enhances algorithm stability, and avoids local maxima efficiently.

\begin{figure}[!ht]
  \centering
  \vspace{-2.7mm}
  \subfigure[]{\includegraphics[width=0.53\linewidth]{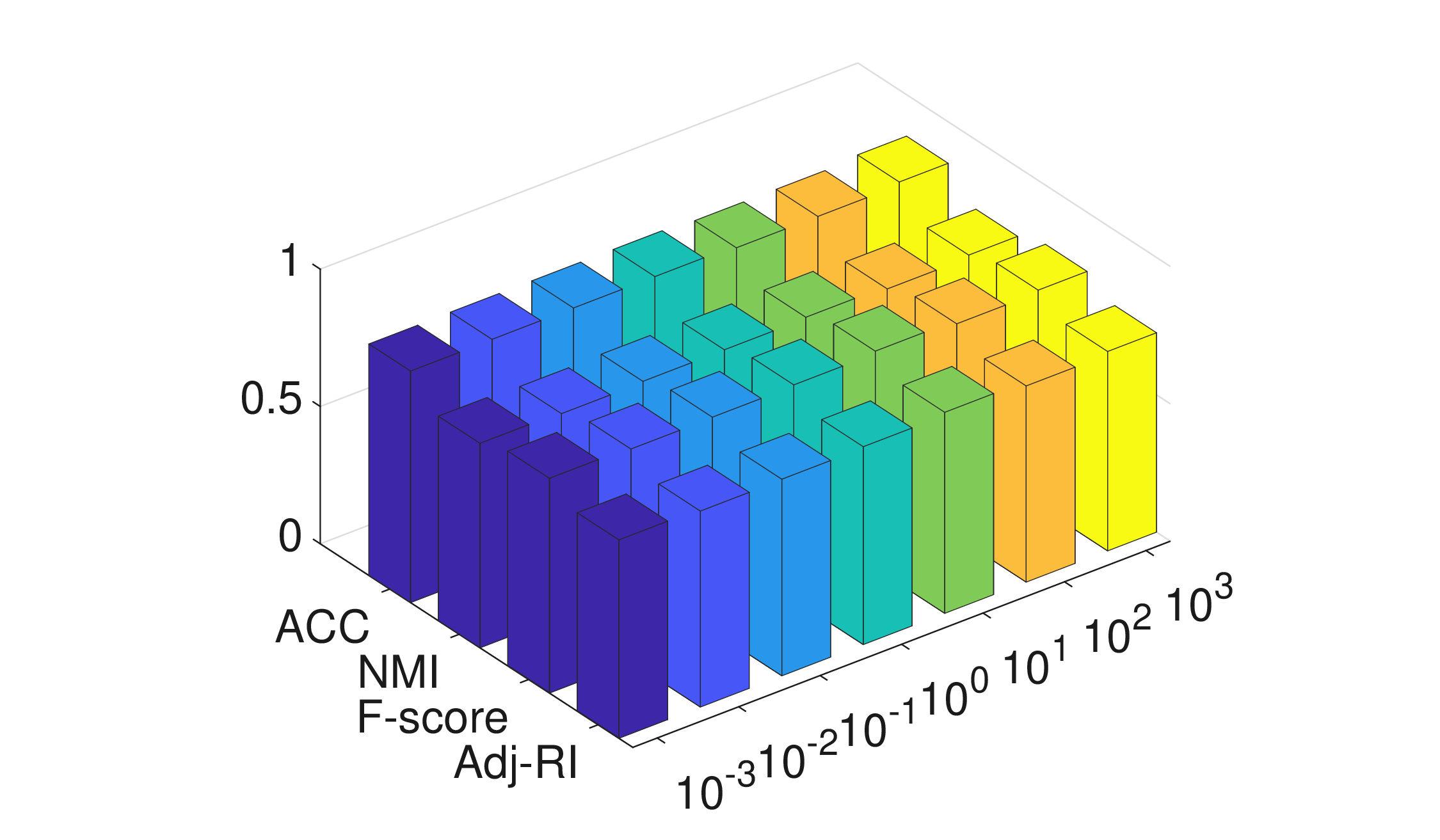}}\hspace{-7mm}
  \subfigure[]{\includegraphics[width=0.53\linewidth]{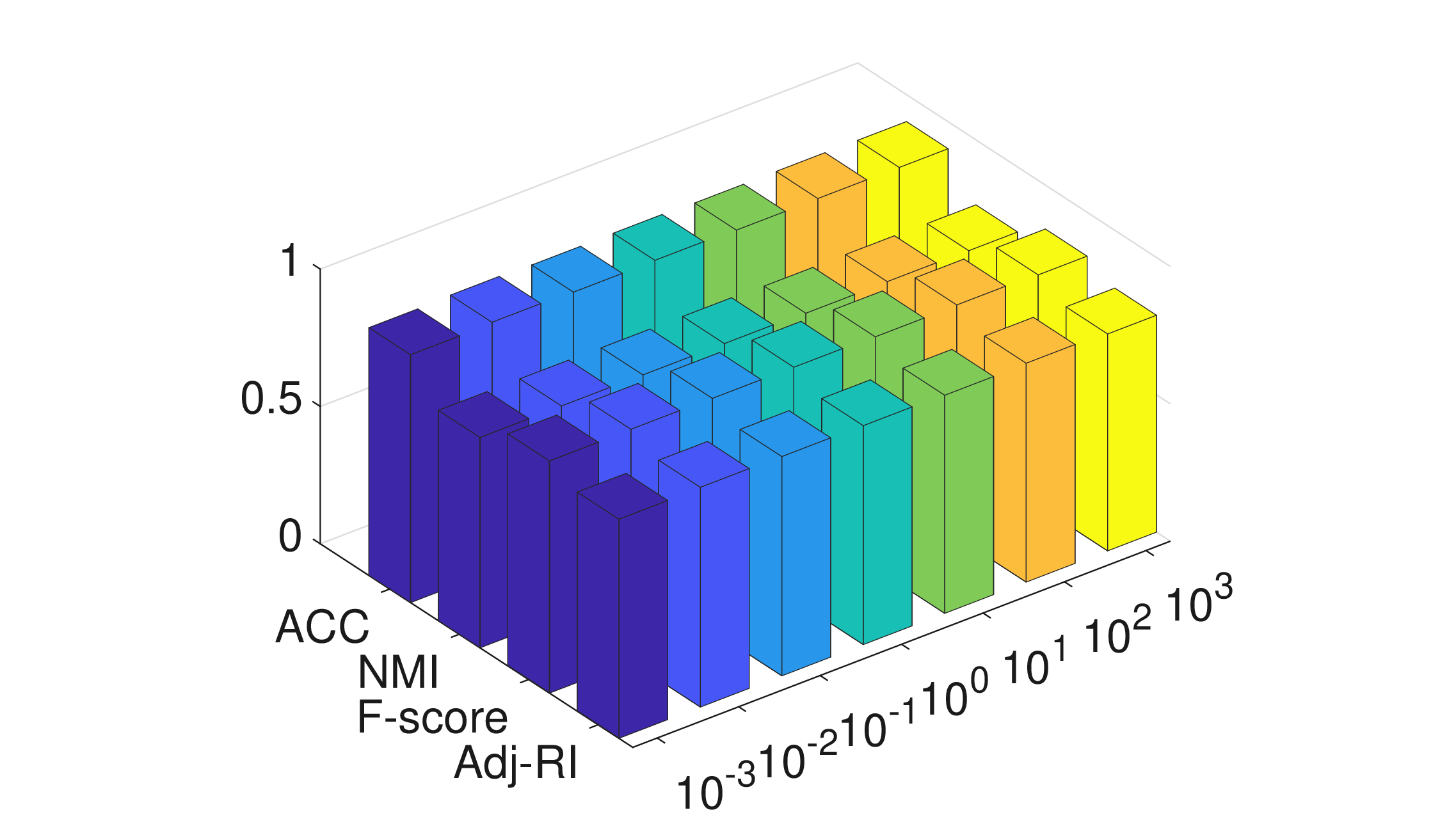}}\\
  \vspace{-2.7mm}
  \subfigure[]{\includegraphics[width=0.53\linewidth]{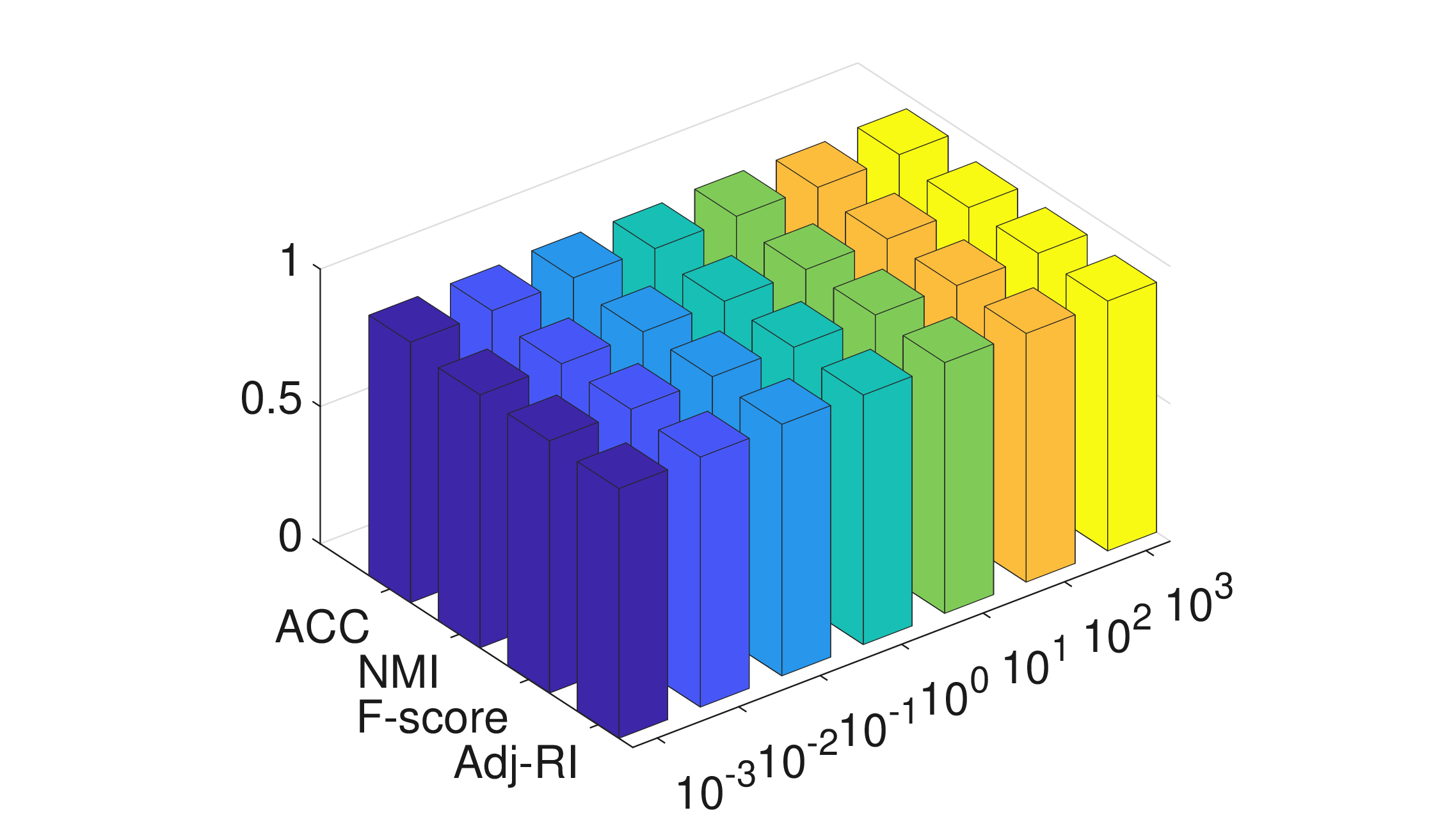}}\hspace{-7mm}
  \subfigure[]{\includegraphics[width=0.53\linewidth]{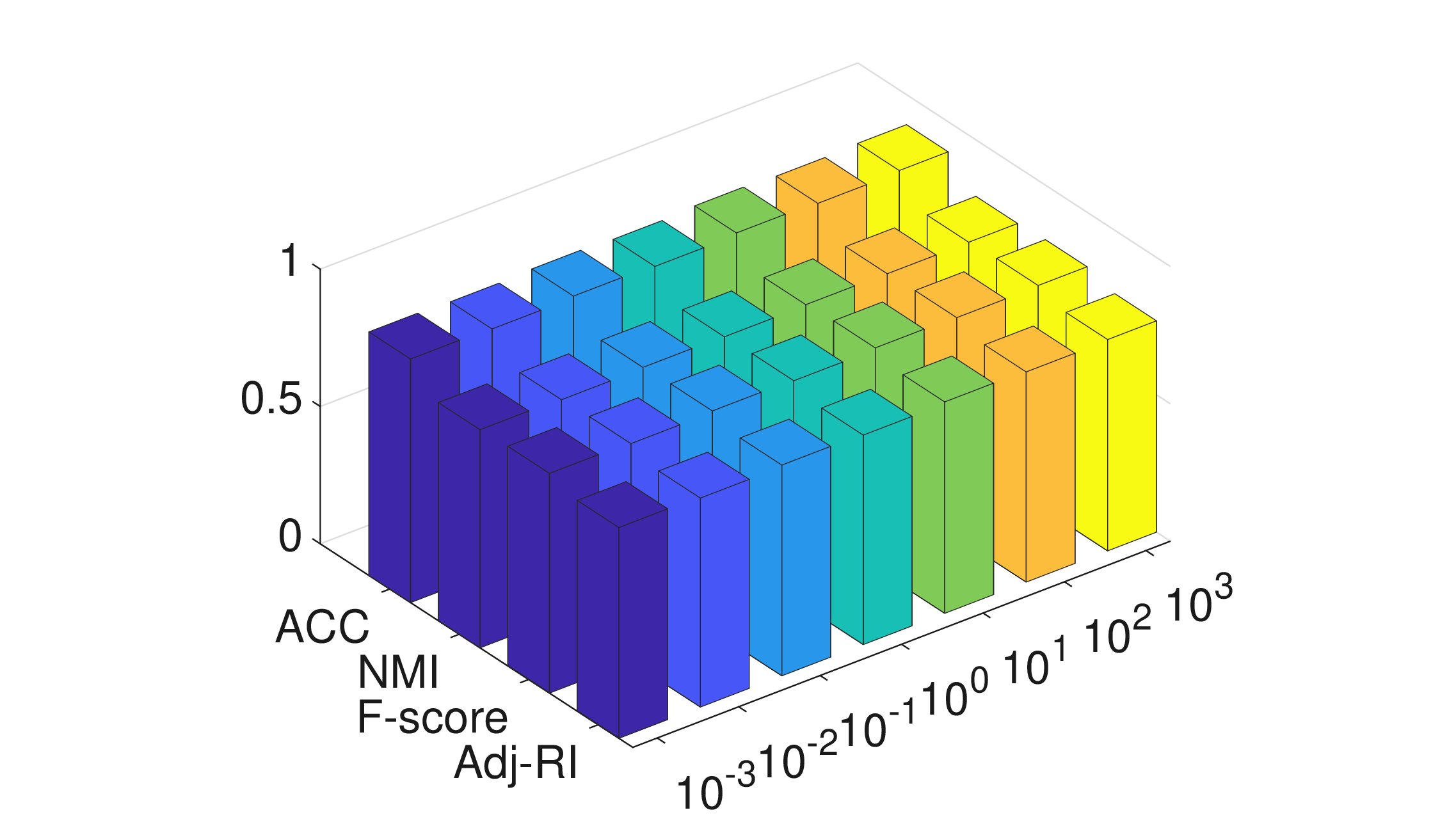}}\\
  \caption{Clustering accuracy of MHSCG with different initial values of $\lambda_l$. (a) 3sources. (b) BBC. (c) Mfeat Digits. (d) MSRCv1. }
  \label{fig:zhu}
\end{figure}

\begin{figure}[!ht]
\centering	
\vspace{-2.2mm}
\subfigure[]{
\includegraphics[width=0.46\linewidth]{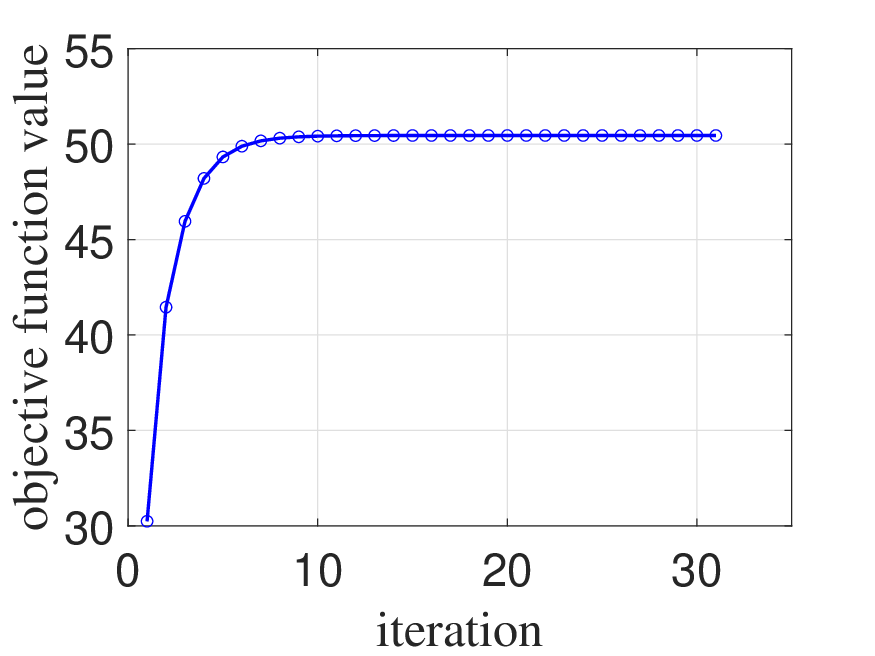}}
\subfigure[]{
\includegraphics[width=0.46\linewidth]{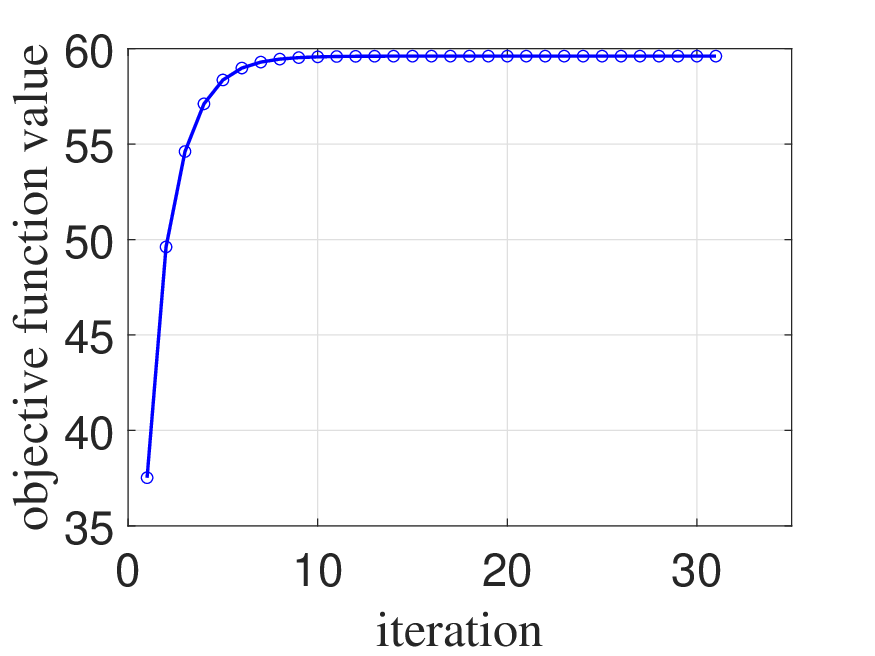}}\\
\vspace{-2.2mm}
 \subfigure[]{
  \includegraphics[width=0.46\linewidth]{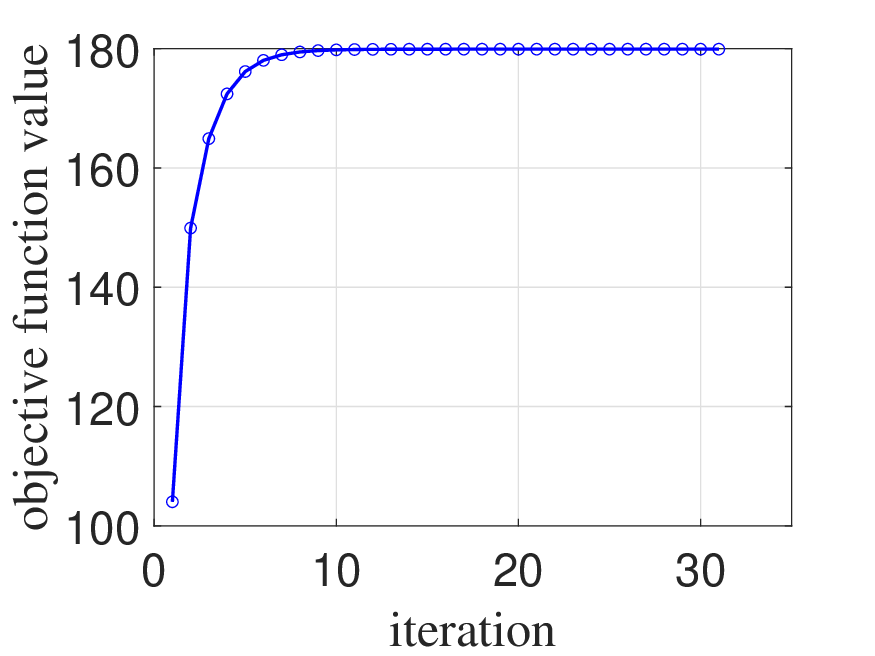}}
   \subfigure[]{
   \includegraphics[width=0.46\linewidth]{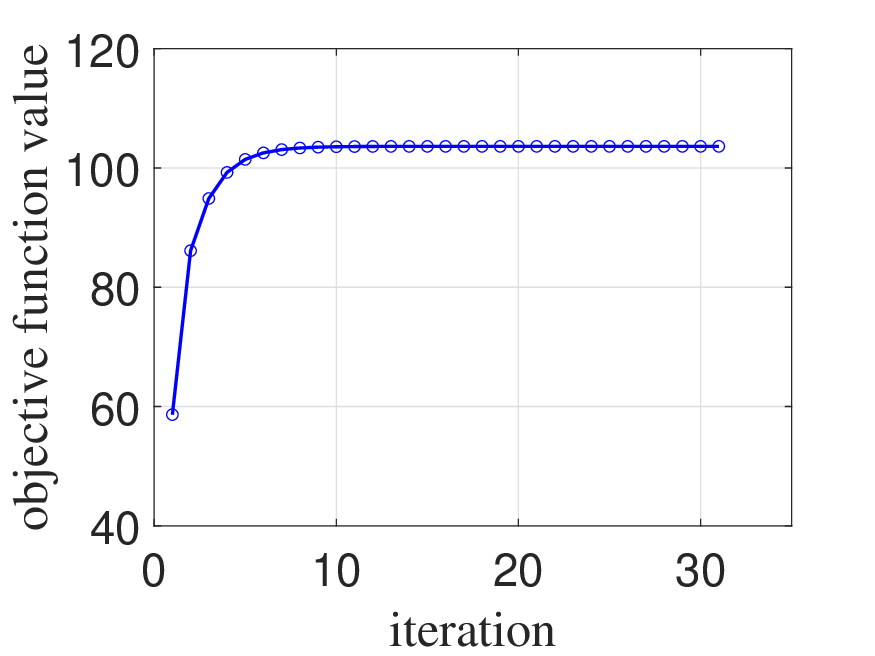}}\\
\caption{Convergence curves of MHSCG. (a) 3sources. (b) BBC. (c) Mfeat Digits. (d) MSRCv1. }
	\label{fig:E2}
\end{figure}

In Algorithm \ref{alg2}, the weighted parameter $\lambda_l$ is adaptively updated, with the initial values for each view set equal, $\lambda_1=\cdots=\lambda_r=\lambda$, ranging from $10^{-3}$ to $10^{3}$. In Fig. \ref{fig:zhu}, we explore the impact of the initial values of $\lambda_l$ on clustering accuracy. Apparently, all clustering metrics remain almost unchanged with different initial values, highlighting the robustness of the proposed algorithm to the initial value of $\lambda_l$.

Finally, we present the convergence curves of the proposed algorithm in Fig. \ref{fig:E2}. Consistent with the theoretical analysis in Section \ref{Convergence}, all curves demonstrate that the objective function's value tends to stabilize at a certain maximum after a few iterations, thereby confirming the algorithm's efficient and stable convergence.

\subsection{Statistical Comparisons}

\begin{figure}
  \centering
  \includegraphics[width=0.47\textwidth]{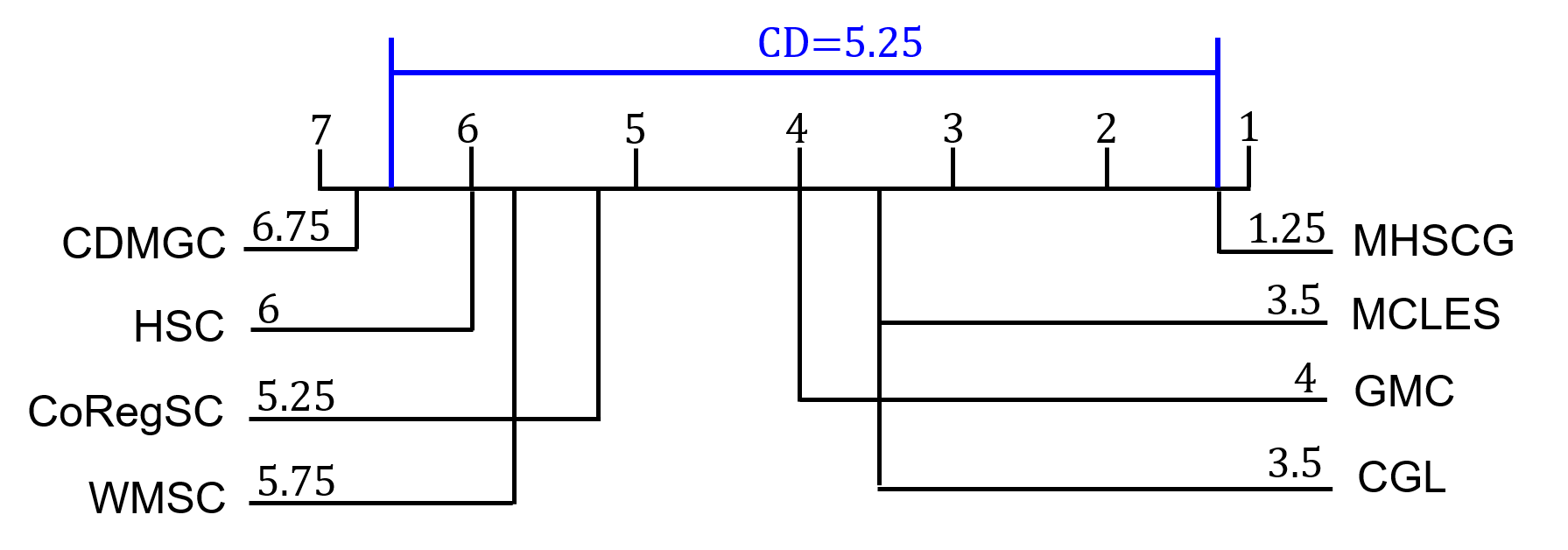}\\
\caption{Comparison of MHSCG with seven baselines.}
\label{fig44}
\end{figure}

To evaluate significant performance differences between the proposed method and seven baselines in multi-view clustering task, we perform statistical comparisons using the Friedman test and the Nemenyi post-hoc test \cite{2006Statistical}. We start by setting the null hypothesis that all algorithms perform equally well on the multi-view clustering task. We denote the number of datasets as $n_d = 4$ and the number of algorithms as $n_k = 8$. We further introduce $R_j$ to represent the average rank of the $j$-th algorithm with $j=1,\ldots,n_k$. The standard Friedman statistic is calculated as
$$
\chi_F^2=\frac{12 n_d}{n_k(n_k+1)}\left[\sum_{j=1}^k R_j^2-\frac{n_k(n_k+1)^2}{4}\right]=14.83.
$$
Because $\chi_F^2$ is undesirably conservative, we adopt a frequently utilized alternative:
$$
F_F=\frac{(n_d-1) \chi_F^2}{n_d(n_k-1)-\chi_F^2}=3.38.
$$
As $F_F$ exceeds the critical value $2.488$ at a significance level of $\alpha=0.05$, we reject the null hypothesis. Subsequently, we proceed with the Nemenyi post-hoc test for further distinction.  In this test, $q_\alpha=3.031$ when $n_k=8$.
The discrepancy in performance between any two algorithms can be assessed by comparing their respective average rank differences, with the critical difference (CD) value calculated as
$$
C D=q_\alpha \sqrt{\frac{n_k(n_k+1)}{6 n_d}}=5.25.
$$
As shown in 
Fig. \ref{fig44}, the proposed method with the first average rank demonstrates superior performance compared to all baselines.

\section{Conclusion}\label{s5}

In this paper, we have proposed a novel multi-view clustering framework named MHSCG. First, we generated a hypergraph for each view based on the similarity matrix learned through sparse representation learning. 
Second, we proposed an optimization function for multi-view hypergraph spectral clustering. This function aims at minimizing the ratio of hyperedge cuts in each partition by using the eigenvectors associated with the top $k$ eigenvalues of the Laplacian matrix for each view and minimizing the similarity discrepancy denoted by eigenvectors between different views simultaneously.
Third, we transformed the optimization problem with orthogonality constraints into an unconstrained optimization problem on the Grassmannian manifold. Finally, we devised
an alternating iterative Riemannian optimization algorithm to efficiently address this problem.
To avoid parameter tuning, the proposed algorithm updates the weighted parameter adaptively, which ensures wide applicability and strong robustness to various datasets. Compared to Euclidean frameworks, optimization on the Grassmannian manifold reduces the variable dimensionality and 
and avoids local maxima efficiently.
To test the effectiveness of MHSCG, we have conducted experiments on four multi-view datasets: 3sources, BBC, 
Mfeat Digits, and MSRCv1.
We compared MHSCG with seven  state-of-the-art multi-view clustering algorithms: HSC, CoRegSC, WMSC, GMC, CGL, MCLES, and CDMGC. The experimental results demonstrate that MHSCG outperforms seven baselines in terms of four clustering evaluation metrics.

While the proposed method introduces innovative contributions, it also faces several challenges.
First, the computational cost of operations is high for large-scale datasets, such as spectral decomposition of Laplacian matrix on the entire hypergraph structure. Second, we adopt the $L_2$-norm in Euclidean space to learn the similarity matrix and generate hypergraphs. There is no unified standard for implicit hypergraph generation methods, as they are often data-driven.
Third, we utilize consistent information for multi-view clustering. It is important to recognize that some multi-view datasets present not only consistency but also inconsistency across various views. Incorporating both consistency and inconsistency is vital for effectively modeling such datasets.

\bibliographystyle{IEEEtran}
\bibliography{IEEEabrv,ref}

\begin{thebibliography}{10}
\providecommand{\url}[1]{#1}
\csname url@samestyle\endcsname
\providecommand{\newblock}{\relax}
\providecommand{\bibinfo}[2]{#2}
\providecommand{\BIBentrySTDinterwordspacing}{\spaceskip=0pt\relax}
\providecommand{\BIBentryALTinterwordstretchfactor}{4}
\providecommand{\BIBentryALTinterwordspacing}{\spaceskip=\fontdimen2\font plus
\BIBentryALTinterwordstretchfactor\fontdimen3\font minus
  \fontdimen4\font\relax}
\providecommand{\BIBforeignlanguage}[2]{{%
\expandafter\ifx\csname l@#1\endcsname\relax
\typeout{** WARNING: IEEEtran.bst: No hyphenation pattern has been}%
\typeout{** loaded for the language `#1'. Using the pattern for}%
\typeout{** the default language instead.}%
\else
\language=\csname l@#1\endcsname
\fi
#2}}
\providecommand{\BIBdecl}{\relax}
\BIBdecl

\bibitem{8731740}
F.~Nie, L.~Tian, R.~Wang, and X.~Li, ``Multiview semi-supervised learning model
  for image classification,'' \emph{IEEE Trans. Knowl. Data Eng.}, vol.~32,
  no.~12, pp. 2389--2400, 2020.

\bibitem{8502831}
C.~Zhang, H.~Fu, Q.~Hu, X.~Cao, Y.~Xie, D.~Tao, and D.~Xu, ``Generalized latent
  multi-view subspace clustering,'' \emph{IEEE Trans. Pattern Anal. Mach.
  Intell.}, vol.~42, no.~1, pp. 86--99, 2020.

\bibitem{9395530}
G.~Chao, S.~Sun, and J.~Bi, ``A survey on multiview clustering,'' \emph{IEEE
  Trans. Artif. Intell.}, vol.~2, no.~2, pp. 146--168, 2021.

\bibitem{8662703}
H.~Wang, Y.~Yang, and B.~Liu, ``{GMC}: Graph-based multi-view clustering,''
  \emph{IEEE Trans. Knowl. Data Eng.}, vol.~32, no.~6, pp. 1116--1129, 2020.

\bibitem{8970909}
Y.~Liang, D.~Huang, and C.~Wang, ``Consistency meets inconsistency: A unified
  graph learning framework for multi-view clustering,'' in \emph{Proc. IEEE
  Int. Conf. Data Min.}, 2019, pp. 1204--1209.

\bibitem{2020MultiChen}
M.~S. Chen, L.~Huang, C.~D. Wang, and D.~Huang, ``Multi-view clustering in
  latent embedding space,'' in \emph{Proc. AAAI Conf. Artif. Intell.}, vol.~34,
  no.~4, 2020, pp. 3513--3520.

\bibitem{2021Measuring}
S.~Huang, I.~Tsang, Z.~Xu, and J.~C. Lv, ``Measuring diversity in graph
  learning: A unified framework for structured multi-view clustering,''
  \emph{IEEE Trans. Knowl. Data Eng.}, vol.~PP, no.~99, pp. 1--14, 2021.

\bibitem{NIPS2011_31839b03}
A.~Kumar, P.~Rai, and H.~Daume, ``Co-regularized multi-view spectral
  clustering,'' in \emph{Proc. Adv. Neural Inf. Process. Syst.}, vol.~24, 2011,
  pp. 1413--1421.

\bibitem{Zong}
L.~Zong, X.~Zhang, X.~Liu, and H.~Yu, ``Weighted multi-view spectral clustering
  based on spectral perturbation,'' in \emph{Proc. AAAI Conf. Artif. Intell.},
  2016, pp. 1969--1976.

\bibitem{2006Learning}
D.~Zhou, J.~Huang, and B.~Schlkopf, ``Learning with hypergraphs: clustering,
  classification, and embedding,'' in \emph{Proc. Adv. Neural Inf. Process.
  Syst.}, 2006, pp. 4--7.

\bibitem{9535255}
X.~Ma, W.~Liu, Q.~Tian, and Y.~Gao, ``Learning representation on optimized
  high-order manifold for visual classification,'' \emph{IEEE Trans.
  Multimedia}, vol.~24, pp. 3989--4001, 2022.

\bibitem{10336546}
Z.~Lin, Q.~Yan, W.~Liu, S.~Wang, M.~Wang, Y.~Tan, and C.~Yang, ``Automatic
  hypergraph generation for enhancing recommendation with sparse
  optimization,'' \emph{IEEE Trans. Multimedia}, vol.~26, pp. 5680--5693, 2024.

\bibitem{6200340}
Y.~Gao, M.~Wang, D.~Tao, R.~Ji, and Q.~Dai, ``{3-D} object retrieval and
  recognition with hypergraph analysis,'' \emph{IEEE Trans. Image Process.},
  vol.~21, no.~9, pp. 4290--4303, 2012.

\bibitem{7993002}
M.~Liu, Y.~Gao, P.-T. Yap, and D.~Shen, ``Multi-hypergraph learning for
  incomplete multimodality data,'' \emph{IEEE J. Biomed. Health Inform.},
  vol.~22, no.~4, pp. 1197--1208, 2018.

\bibitem{Huang_2015_CVPR}
S.~Huang, M.~Elhoseiny, A.~Elgammal, and D.~Yang, ``Learning
  hypergraph-regularized attribute predictors,'' in \emph{Proc. IEEE conf.
  comput. vis. pattern recognit.}, pp. 409--417.

\bibitem{10.1145/3308558.3313635}
D.~Yang, B.~Qu, J.~Yang, and P.~Cudre-Mauroux, ``Revisiting user mobility and
  social relationships in lbsns: A hypergraph embedding approach,'' in
  \emph{WWW Conf.}, 2019, pp. 2147--2157.

\bibitem{7064739}
M.~Wang, X.~Liu, and X.~Wu, ``Visual classification by $\ell _1$-hypergraph
  modeling,'' \emph{IEEE Trans. Knowl. Data Eng.}, vol.~27, no.~9, pp.
  2564--2574, 2015.

\bibitem{2016Elastic}
Q.~Liu, Y.~Sun, C.~Wang, T.~Liu, and D.~Tao, ``Elastic net hypergraph learning
  for image clustering and semi-supervised classification,'' \emph{IEEE Trans.
  Image Process.}, vol.~26, no.~1, pp. 452--463, 2017.

\bibitem{2019Robust}
T.~Jin, Z.~Yu, Y.~Gao, S.~Gao, X.~Sun, and C.~Li, ``Robust $\ell_2$-hypergraph
  and its applications,'' \emph{Inf. Sci.}, vol. 501, pp. 708--723, 2019.

\bibitem{8590732}
T.~Jin, R.~Ji, Y.~Gao, X.~Sun, X.~Zhao, and D.~Tao, ``Correntropy-induced
  robust low-rank hypergraph,'' \emph{IEEE Trans. Image Process.}, vol.~28,
  no.~6, pp. 2755--2769, 2019.

\bibitem{9859633}
Y.~Hu and H.~Cai, ``Multi-view clustering through hypergraphs integration on
  stiefel manifold,'' in \emph{IEEE Int. Conf. Multimed. Expo (ICME)}, 2022,
  pp. 01--06.

\bibitem{8099818}
Q.~Wang, J.~Gao, and H.~Li, ``Grassmannian manifold optimization assisted
  sparse spectral clustering,'' in \emph{Proc. IEEE Conf. Comput. Vis. Pattern
  Recognit.}, 2017, pp. 3145--3153.

\bibitem{Pasadakis}
D.~Pasadakis, C.~L. Alappat, O.~Schenk, and G.~Wellein, ``Multiway p-spectral
  graph cuts on grassmann manifolds,'' \emph{Mach. Learn.}, vol. 111, no.~2,
  pp. 791--829, 2021.

\bibitem{2011Multiview}
T.~Xia, D.~Tao, M.~Tao, and Y.~Zhang, ``Multiview spectral embedding,''
  \emph{IEEE Trans. Cybern.}, vol.~40, no.~6, pp. 1438--1446, 2011.

\bibitem{2018Multiview}
Y.~Wang, L.~Wu, X.~Lin, and J.~Gao, ``Multiview spectral clustering via
  structured low-rank matrix factorization,'' \emph{IEEE Trans. Neural Netw.
  Learn. Syst.}, vol.~29, no.~10, pp. 4833--4843, 2018.

\bibitem{2011A}
\BIBentryALTinterwordspacing
A.~Kumar and H.~Iii, ``A co-training approach for multi-view spectral
  clustering abhishek kumar,'' in \emph{Proc. the 28th Int. Conf. Mach.
  Learn.}, 2011.  \url{http://users.umiacs.umd.edu/~hal/docs/daume11cospec.pdf}
\BIBentrySTDinterwordspacing

\bibitem{2016A}
A.~Appice and D.~Malerba, ``A co-training strategy for multiple view clustering
  in process mining,'' \emph{IEEE Trans. Serv. Comput.}, vol.~9, no.~99, pp.
  832--845, 2016.

\bibitem{6805155}
L.~Huang, J.~Lu, and Y.~Tan, ``Co-learned multi-view spectral clustering for
  face recognition based on image sets,'' \emph{IEEE Signal Process. Lett.},
  vol.~21, no.~7, pp. 875--879, 2014.

\bibitem{7451227}
C.~Lu, S.~Yan, and Z.~Lin, ``Convex sparse spectral clustering: Single-view to
  multi-view,'' \emph{IEEE Trans. Image Process.}, vol.~25, no.~6, pp.
  2833--2843, 2016.

\bibitem{2010Image}
Y.~Huang, Q.~Liu, S.~Zhang, and D.~N. Metaxas, ``Image retrieval via
  probabilistic hypergraph ranking,'' in \emph{Proc. IEEE Conf. Comput. Vis.
  Pattern Recognit.}, 2010, pp. 3376--3383.

\bibitem{2018Learning}
W.~Zhao, S.~Tan, Z.~Guan, B.~Zhang, M.~Gong, Z.~Cao, and Q.~Wang, ``Learning to
  map social network users by unified manifold alignment on hypergraph,''
  \emph{IEEE Trans. Neural Netw. Learn. Syst.}, vol.~29, no.~12, pp.
  5834--5846, 2018.

\bibitem{Xueqi2018Hypergraph}
X.~Ma, W.~Liu, S.~Li, D.~Tao, and Y.~Zhou, ``Hypergraph $p$-laplacian
  regularization for remotely sensed image recognition,'' \emph{IEEE Trans.
  Geosci. Remote Sensing}, vol.~57, no.~3, pp. 1585--1595, 2019.

\bibitem{2019Random}
\BIBentryALTinterwordspacing
T.~Carletti, F.~Battiston, G.~Cencetti, and D.~Fanelli, ``Random walks on
  hypergraphs,'' \emph{Phys. Rev. E}, vol. 101, no.~2, 2019.
  \url{https://doi.org/10.1103/physreve.101.022308}
\BIBentrySTDinterwordspacing

\bibitem{2019HypergraphFeng}
Y.~Feng, H.~You, Z.~Zhang, R.~Ji, and Y.~Gao, ``Hypergraph neural networks,''
  \emph{Proc. AAAI Conf. Artif. Intell.}, vol.~33, pp. 3558--3565, 2019.

\bibitem{2021Hypergraph}
\BIBentryALTinterwordspacing
S.~Bai, F.~Zhang, and P.~Torr, ``Hypergraph convolution and hypergraph
  attention,'' \emph{Pattern Recognit.}, vol. 110, 2021.
  \url{http://dx.doi.org/10.1016/j.patcog.2020.107637}
\BIBentrySTDinterwordspacing

\bibitem{1998The}
A.~Edelman, T.~A., and S.~Smith, ``The geometry of algorithms with
  orthogonality constraints,'' \emph{SIAM J. Matrix Anal. Appl.}, vol.~20,
  no.~2, pp. 303--353, 1998.

\bibitem{2008Grassmann}
J.~Ham and D.~Lee, ``Grassmann discriminant analysis: a unifying view on
  subspace-based learning,'' in \emph{Proc. 25th Int. Conf. Mach. Learn.},
  2008, pp. 376--383.

\bibitem{10.5555/3016100.3016174}
F.~Nie, X.~Wang, M.~Jordan, and H.~Huang, ``The constrained laplacian rank
  algorithm for graph-based clustering,'' in \emph{Proc. AAAI Conf. Artif.
  Intell.}, 2016, pp. 1969--1976.

\bibitem{Mishra}
B.~Mishra, H.~Kasai, P.~Jawanpuria, and A.~Saroop, ``A riemannian gossip
  approach to subspace learning on grassmann manifold,'' \emph{Mach. Learn.},
  vol. 108, no.~10, pp. 1783--1803, 2019.

\bibitem{2006Statistical}
J.~Demiar and D.~Schuurmans, ``Statistical comparisons of classifiers over
  multiple data sets,'' \emph{J. Mach. Learn. Res.}, vol.~7, no.~1, pp. 1--30,
  2006.

\end{thebibliography}

\vfill

\end{document}